\documentclass{article}

     \usepackage[preprint]{neurips_2023}

\usepackage[utf8]{inputenc} 
\usepackage[T1]{fontenc}    
\usepackage{hyperref}       
\usepackage{url}            
\usepackage{booktabs}       
\usepackage{amsfonts}       
\usepackage{amsmath}       
\usepackage{nicefrac}       
\usepackage{microtype}      
\usepackage{xcolor}         

\usepackage{multibib}
\newcites{SM}{Supplementary References}

\usepackage[pdftex]{graphicx}
\usepackage{multirow}

\title{Sparsity-depth Tradeoff in Infinitely Wide Deep Neural Networks}

\author{%
  Chanwoo Chun \\
  Weill Cornell Medical College of Cornell University\\
  New York, NY 10021 \\
  \texttt{cc2465@cornell.edu} \\
  \And
  Daniel D. Lee \\
  Cornell Tech\\
  New York, NY 10044\\
  \texttt{ddl46@cornell.edu} \\
}

\begin{document}

\maketitle
\begin{abstract}
We investigate how sparse neural activity affects the generalization performance of a deep Bayesian neural network at the large width limit. To this end, we derive a neural network Gaussian Process (NNGP) kernel with rectified linear unit (ReLU) activation and a predetermined fraction of active neurons. Using the NNGP kernel, we observe that the sparser networks outperform the non-sparse networks at shallow depths on a variety of datasets. We validate this observation by extending the existing theory on the generalization error of kernel-ridge regression.
\end{abstract}

\section{Introduction}
The utility of sparse neural representations has been of interest in both the machine learning and neuroscience communities. \citet{willshaw1990optimal} first showed that sparse inputs accelerate learning in a single hidden layer neural network. More recently, \citet{babadi2014sparseness} analyzed how sparse expansion of a random single hidden layer network modeling the cerebellum enhances the classification performance by reducing both the intraclass variability and excess overlaps between classes. In this work, we examine the effect of sparsity on the generalization performance for regression and regression-based classification tasks in deeper neural networks with rectified linear activations.

\begin{figure}
    \centering
    \includegraphics[width=1\linewidth]{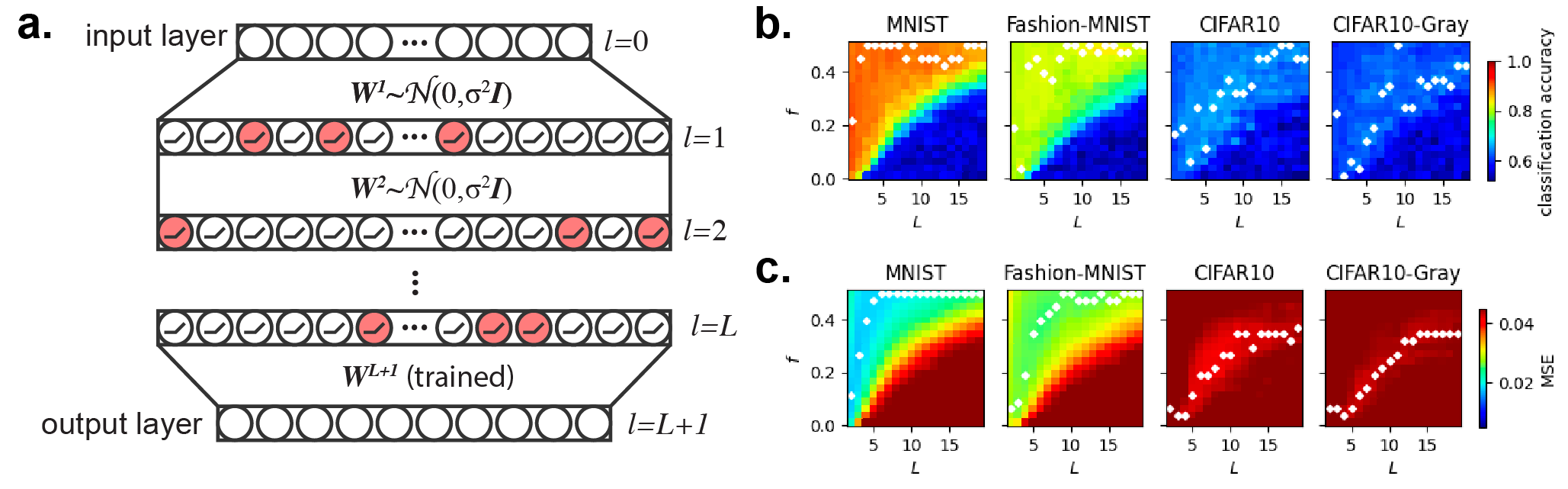}
    \caption{(a) Sparse and deep neural network with random intermediate weights and trained last output layer. The intermediate layer neurons are rectified linear units (ReLU). A fixed fraction ($f$) of neurons in each layer are non-zero (red neurons) for a given input; (b,c) Numerical simulation of very wide sparse neural networks over a range of sparsity $f$ and depth $L$. Each layer contains 20,000 neurons and the outputs are learned using 100 training examples. (b) Classification accuracy of the models as depth and sparsity are varied. The best-performing model of each depth is indicated with a white marker; (c) Mean-square error (MSE) of the regressions. The model with minimum MSE solution is indicated with a white marker for each depth.}
    \label{finite}
\end{figure}

Consider a feed-forward deep neural network with a large number of neurons equipped with rectified linear units (ReLU) in each layer (Figure \ref{finite}a). The weights are random and, for each input, we adjust the bias in the preactivations such that only a fraction $f$ neurons in each layer are positive after ReLU. This sparse random network is trained by optimally tuning the readout layer using the pseudo-inverse rule. We performed regressions on the one-hot vectors of the real-life datasets, i.e. MNIST, Fashion-MNIST, CIFAR10, and CIFAR10-Grayscale, with 100 training samples \citep{lecun2010,fashion,cifar}. Interestingly, the sparsity of the model with the best generalization performance changes over depths (Figure \ref{finite}b,c). At shallow depth, the sparse activation improves the generalization performance, whereas at the deeper configurations, denser activations are required to maintain high generalization performance. 
To theoretically analyze the performance of these networks, we will take the width of the intermediate layers to be very large. It has been well-established that Bayesian inference on an infinite width feedforward neural network is equivalent to training only the readout weights of the network \citep{matthews2017sample,hron2020exact,hron2022wide,williams1996computing,gb2018}. This allows us to perform kernel analysis of infinite-width neural networks assuming normally distributed weights, and to exactly infer the posterior network output via kernel ridge regression. Such kernels are referred to as neural network Gaussian process (NNGP) kernels. \citet{cho2009kernel} introduced the ReLU neural network kernel which has been experimentally shown by \citet{gb2018} to have a performance comparable to finite neural networks learned with backpropagation. \citet{gb2018} performed regression on the one-hot training vectors and took the max of the prediction vector to obtain the test accuracy, while also reporting the mean-squared error of the regression. In another work by \citet{cho2011analysis}, they derived an NNGP with Heaviside step activation to induce sparse activation. Here we present a deep NNGP with sparse activation induced by ReLU and appropriately chosen biases, and investigate its generalization performance 
as shown from the numerical experiments in Figure \ref{finite}.

To better understand the generalization performance of these networks, we employ the theoretical framework provided in \citet{canatar2021out,canatar2021spectral}. \citet{canatar2021spectral} used the replica method to derive an analytical expression for the in-distribution generalization error for kernel ridge regression. There exists a large body of literature on the generalization bound and its convergence rate of kernel ridge or ridgeless regression \citep{spigler2020asymptotic,bordelon2020spectrum,bietti2020deep,scetbon2021spectral,vakili2021uniform}. In our paper, however, the goal is to characterize the average generalization performance of a kernel. By averaging over the data distribution,\citet{bordelon2020spectrum} and \citet{canatar2021spectral} formulate the average generalization error as an analytical function of the kernel spectrum and the target function spectrum. We show that their theoretical formula accurately matches our experimental observation, and extend the theory to allow intuitive comparisons between kernels.

\subsection{Summary of contributions}
We begin by deriving the expression for the sparse NNGP kernel in Section \ref{deriv}. We then experimentally demonstrate in Section \ref{experi} that the sparse NNGP $f<0.5$ outperforms the popular NNGP kernel, i.e. ReLU arccosine kernel $f=0.5$, at shallow depths. The arccosine kernel without bias is a special case of our sparse NNGP kernel where the fraction of the active neurons is $f=0.5$. The bias typically does not affect generalization performance (see Supp \ref{app:bnoise} and \citet{gb2018}).

Next, in Section \ref{theorySec} we expand on the existing theory for kernel ridge-regression provided in \citet{canatar2021spectral} to aid our understanding of the generalization performance of the sparse NNGP. Our theoretical contribution is showing the intuitive relationship between the shape of the kernel eigenspectrum and the shape of the modal error spectrum using first-order perturbation theory, which provides useful insight when comparing kernel functions. 

\section{Sparse neural network Gaussian process}\label{deriv}
\subsection{Architecture}
We consider a fully connected feed-forward neural network architecture. The post-activation $x^{l}$ of each neuron in layer $l$ is a rectified
version of a preactivation $h^{l}$ shifted by a bias $b^{l}$. The
preactivation itself is a linear combination of the previous layer
activity $x^{l-1}$. The model is written as
\begin{equation}\label{finite0}
x_{j}^{(p),l}=\left[h_{j}^{(p),l}-b^{(p),l}\right]_{+}    
\quad\quad
h_{j}^{(p),l}=\sum_{i=1}^{n_{l-1}}w_{ij}^{l}x_{i}^{(p),l-1}    
\end{equation}
For the final output $h_{j}^{(p),L+1}$ is
\begin{equation}\label{eq:output}
h_{j}^{(p),L+1}=\sum_{i=1}^{n_L}w_{ij}^{L+1}x_{i}^{(p),L}    
\end{equation}
$w^l_{ij}$ denotes a synaptic weight from neuron $i$ of layer $l-1$
to neuron $j$ of layer $l$. In each layer, there is $n_{l}$ number
of neurons. The superscript $(p),l$ denotes the input sample index and layer
number respectively. For the input, we drop the $l$ superscript, i.e. $x_{i}^{(p)}$ instead of $x_{i}^{(p),0}$.  The output layer $L+1$ neuron does not have an activation function, so $\{h_0^{(p),L+1} \ldots h_{n^{L-1}}^{(p),L+1}\}$ is considered the model output. $L$ denotes the number of hidden layers.

For each forward pass of an input, the biases are adjusted such that a fixed fraction $f$ of neurons are positive in each layer. After the rectification by ReLU, only the $f$ fraction of neurons are non-zero.


We take the infinite width limit to enable exact Bayesian inference. To this end, we first derive the sparse NNGP kernel formula for a single hidden layer architecture and then compose it to arrive at the sparse NNGP for deep architectures.

\subsection{Prior of the single hidden layer architecture ($L=1$)}
For the prior, the weights are independently sampled from a zero-mean normal distribution with standard deviation $\frac{\sigma}{\sqrt{n^{l-1}}}$, i.e. $w_{ij}^l \sim \mathcal{N}\left(0,\frac{\sigma^2}{n^{l-1}}\right)$ for $l=1$. Since the inputs $x^{(p)}_k$ are fixed, the preactivation of the hidden layer $h_{i}^{(p)}$, which is the sum of the inputs weighted by the normal random weights, is a zero-mean normal with standard deviation $\sigma_h = \frac{\sigma}{\sqrt{n}}\|x^{(p)}\|$, where $\|x^{(p)}\|=\sqrt{\sum_{k=1}^{n} \left[x_k^{(p)}\right]^2}$ and $n$ is the dimension of the input. In other words, $h_i^{(p)}\sim \mathcal{N}\left(0,\frac{\sigma^2}{n}\|x^{(p)}\|^2 \right)$. Since we know the preactivation is normally distributed, the thresholded rectification $\left[ h^{(p)}_i - b^{(p)} \right]_+$ of it is a rectified normal distribution. Since we want a fixed level of sparsity, we require a predetermined fraction $f$ of the rectified normal distribution to be positive, i.e. non-zero, by choosing the appropriate bias. If $\sigma_h^{2}$ is the variance of $h_{i}^{(p)}$, the bias that guarantees exactly $f$ fraction of the neurons to be positive is $b^{(p)}=\sigma_{h}\tau$ where \(\tau=\sqrt{2}\text{erf}^{-1}(1-2f)\). Note that $b^{(p)}$ is a function of the input, since it is dependent on $\sigma_{h}$ which is a function of the input norm.

For finite $n^{l=1}$, the output $h_j^{(p),l=2}$ is non-normal, since it is a dot product between normal random weights $ w_{ij}^{l=2} $ and rectified normal activities $x_i^{(p),1}$ of the hidden layer (Eqn. \ref{eq:output}). However, when $n^{l=1}\rightarrow \infty$, we can invoke the central limit theorem, and the distribution of $h_j^{(p),l=2}$ reaches a normal distribution \citep{neal1996priors}.

In order to compute the posterior output of this network, we first need to compute the similarity between neural representations of two inputs $p$ and $q$, i.e. $E\left[ x_i^{(p),l=1} x_i^{(q),l=1}\right]$ averaged over the distribution of the weights. This similarity is referred to as the Gaussian process kernel $K(\mathbf{x}^{(p)},\mathbf{x}^{(q)})$, where $\mathbf{x}^{(p)}$ is a vector representation of the input sample $p$. The kernel is computed as
\begin{equation}\label{eq:integral0}
    K(\mathbf{x}^{(p)},\mathbf{x}^{(q)})= \int d\mathbf{w}^{1}_{i} P(\mathbf{w}_{i}^{1}) \times \left[\mathbf{w}_{i}^{1}\cdot\mathbf{x}^{(p)}-b^{(p)}\right]_{+}\left[\mathbf{w}_{i}^{1}\cdot\mathbf{x}^{(q)}-b^{(q)}\right]_{+}
\end{equation}
where $\mathbf{w}_{i}^{1}$ is a vector whose $k^{th}$ element is $w_{ki}^{1}$, a weight between the input and the hidden layers.

The integration (Eqn. \ref{eq:integral0}) can be reduced to a one-dimensional
integration which can be efficiently computed using simple numerical
integration. The resulting formula for the sparse NNGP kernel is
\begin{equation} \label{eq:K0}
    K(\mathbf{x}^{(p)},\mathbf{x}^{(q)}) =\frac{\sigma^{2}}{2\pi}\|\mathbf{x}^{(p)}\| \|\mathbf{x}^{(q)}\| \left(2I\left(\theta\mid\tau\right)- \tau\sqrt{2\pi}(1+\cos\theta)\right)
\end{equation}
\begin{equation} \label{eq:theta0}
    \theta=\arccos\frac{\mathbf{x}^{(p)}\cdot\mathbf{x}^{(q)}}{\|\mathbf{x}^{(p)}\| \|\mathbf{x}^{(q)}\|}
\end{equation}
\begin{multline} \label{eq:I}
    I(\theta\mid\tau)= \int_{0}^{\frac{\pi-\theta}{2}}\exp\left(-\frac{\tau^{2}}{2\sin^{2}(\phi_{0})}\right)2\sin\left(\phi_{0}+\theta\right)\sin(\phi_{0})\\
    +\tau\left(\sin\left(\phi_{0}+\theta\right)+\sin(\phi_{0})\right)\sqrt{\frac{\pi}{2}}\text{erf\ensuremath{\left(\frac{\tau}{\sqrt{2}\sin(\phi_{0})}\right)}\ensuremath{d\ensuremath{\phi_{0}}}}
\end{multline}
Note that $\tau$ is the variable that controls sparsity as defined earlier. As $\tau\rightarrow0$, the kernel is equivalent to the arccosine
kernel of degree 1 and zero bias derived by \citet{cho2009kernel} (see Supp. \ref{app:relate} for the proof). See Supp. \ref{app:NNGP_deriv} for the full derivation of the sparse Kernel.

\subsection{Multi-layered sparse NNGP $L>1$}
In the multilayered formulation of the sparse NNGP, we take all $n^l \rightarrow \infty$. The recursive formula for the multilayered sparse NNGP kernel is
\begin{equation} \label{eq:K}
    K^{l}(\mathbf{x}^{(p)},\mathbf{x}^{(q)}) =\frac{\sigma^{2}}{2\pi}\sqrt{K^{l-1}(\mathbf{x}^{(p)},\mathbf{x}^{(p)})K^{l-1}(\mathbf{x}^{(q)},\mathbf{x}^{(q)})}  \times \left(2I\left(\theta^{l}\mid\tau\right)- \tau\sqrt{2\pi}(1+\cos\theta^{l})\right)
\end{equation}
\begin{equation} \label{eq:theta}
    \theta^{l}=\arccos\frac{K^{l-1}(\mathbf{x}^{(p)},\mathbf{x}^{(q)})}{\sqrt{K^{l-1}(\mathbf{x}^{(p)},\mathbf{x}^{(p)})K^{l-1}(\mathbf{x}^{(q)},\mathbf{x}^{(q)})}}
\end{equation}
See Supp. \ref{app:NNGP_deriv} for the full derivation. This is almost identical to the formulation of the single-layer kernel in Eqn. \ref{eq:K0}-\ref{eq:theta0}, except the dot product, and hence the length are computed differently. In Eqn. \ref{eq:theta0}, the dot product is between the deterministic inputs is $\mathbf{x}^{(p)}\cdot\mathbf{x}^{(q)}$ but in Eqn. \ref{eq:theta} the dot product of the stochastic representations is computed by $K^{l}(\mathbf{x}^{(p)},\mathbf{x}^{(q)})$. Naturally, it follows that the length of a representation is $\sqrt{K^{l}(\mathbf{x}^{(p)},\mathbf{x}^{(p)})}$. The first hidden layer kernel $K^{l=1}(\mathbf{x}^{(p)},\mathbf{x}^{(q)})$ is the same as Eqn. \ref{eq:K0}. Throughout the paper, we assume all layers of a given network have the same sparsity.

\begin{figure}
\centering
\includegraphics[width=1\linewidth]{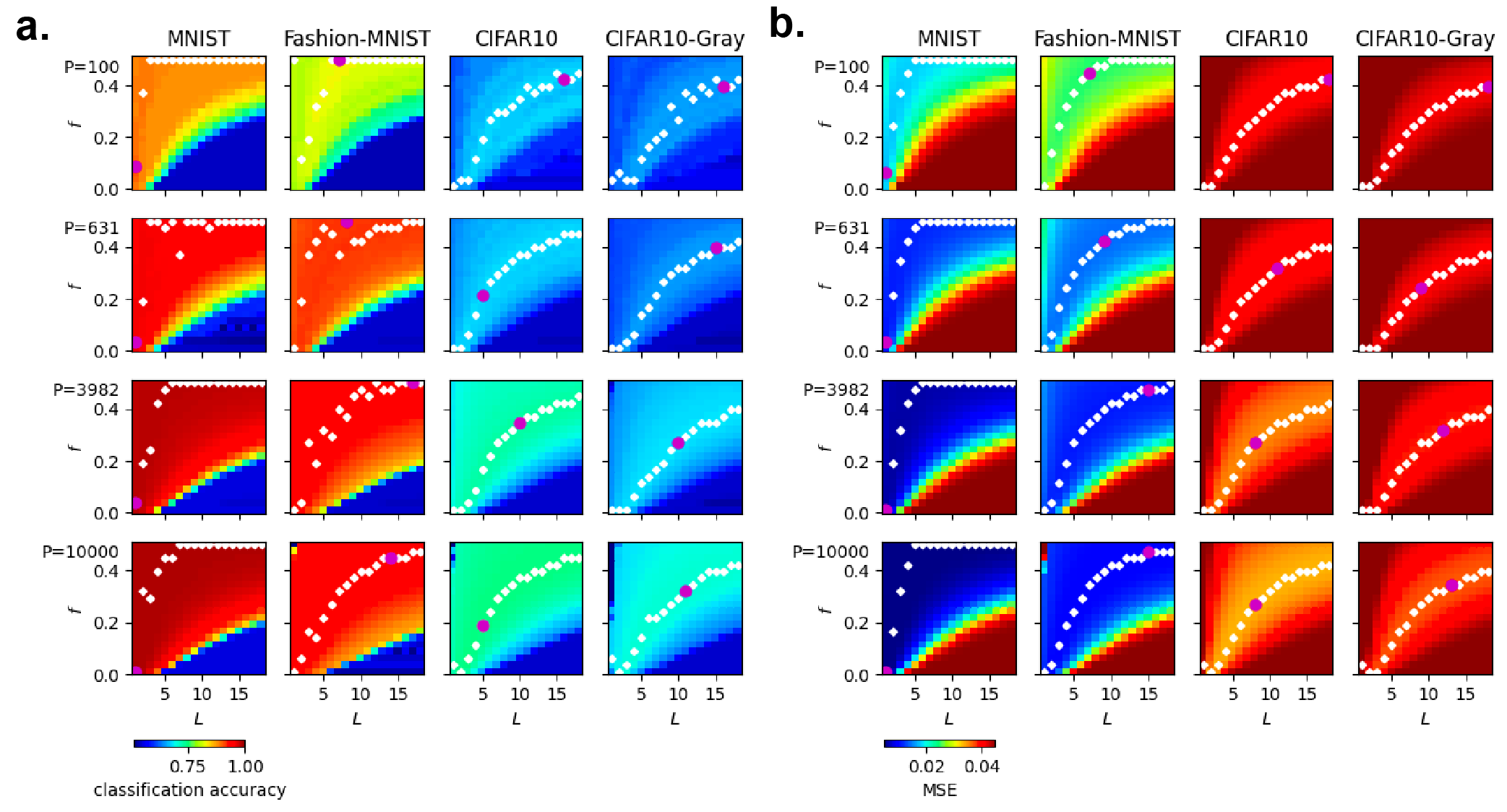}
\caption{Infinite-width Bayesian neural network performance over a range of sparsity and depth. (a) Classification accuracy of the models on real-life datasets. The best-performing model of each depth is indicated with a white marker. Purple dots indicate the best-performing kernel across all sparsities and depths. Each row corresponds to a different number of training samples $P$. (b) Corresponding mean-square error (MSE) of the regressions on the datasets. The model with minimum MSE solution is indicated with a white marker for each depth.}
\label{infinite}
\vspace*{-\intextsep}
\end{figure}

\section{Experimental results on sparse NNGP}\label{experi}
In the infinite-width case, exact Bayesian inference is possible. We directly compute the posterior of the predictive distribution, whose mean ($\mu$) is given by the solution to kernel ridge regression: $\mu=\mathbf{K}^L_*\left(\mathbf{K}^L+\lambda \mathbf{I}\right)^{-1}\mathbf{Y}$
where $\lambda$ is a ridge parameter. $\mathbf{K}^L$ is the kernel gram matrix for the representation similarity within training data at the last hidden layer, and $\mathbf{K}^L_*$ is the kernel gram matrix for the representation similarity between the test and training data. $\mathbf{Y}$ is the training target matrix whose each row is a training sample target and each column is a feature. We train the sparse NNGP on MNIST, Fashion-MNIST, CIFAR10, and grayscale CIFAR10 datasets with different training set sizes (see experimental details in Supp. \ref{app:exp}).

\subsection{Sparsity-depth tradeoff}
As we sweep over the ranges of sparsity $f$ and depth $L$, we see a pattern of the generalization performance over the sparsity vs. depth plane, which we denote the $fL$-plane (Figure \ref{infinite}). The patterns of the generalization performance over the $fL$-plane are more pronounced in the NNGP solutions (Figure \ref{infinite}) compared to the finite-width solutions (Figure \ref{finite}). It is consistent throughout different datasets that sparse networks need to be shallow whereas dense networks need to be deep, in order to gain high performance. When sparse networks are too deep, the performance abruptly drops. The result has a narrow confidence interval over the randomized training sets as shown in Supp. \ref{app:conf}. We also observe a strong preference for sparser models in finite $\lambda>0$ cases (Supp. \ref{app:regular}).

\subsection{Sparse and shallow networks are comparable to dense and deep networks}
The main result of this paper is that the sparse and shallow networks have comparable generalization performances to dense and deep networks. As an example, in the MNIST classification task with $P=1000$, the best-performing dense $f=0.5$ model has depth $L_d=5$ with accuracy $0.9300\pm0.0028$. However, we can find a sparser $f=0.13$ model with shallower depth $L_s=1$ with essentially identical performance $a_s=0.9303\pm0.0025$. At the same depth $L_s=1$, the dense model has an accuracy $a_d=0.9227\pm0.0015$ lower than $a_s$. The observation of $L_s<L_d$ and $a_s>a_d$ is highly consistent throughout different datasets and training sample sizes (see Table \ref{table_result}, and Supp. \ref{app:table}). In short, it is quantitatively clear that a sparse kernel requires a smaller depth, and hence fewer kernel compositions to reach the performance level observed in the deep dense model.

A greater number of kernel compositions requires more computational time. Therefore the shallow and sparse kernel achieves a performance similar to deep and dense $f=0.5$ kernel with less computational time. One may argue that the $f=0.5$ case (arccosine kernel) does not require a numerical integration that is needed for the $f<0.5$ case, so $f=0.5$ kernel is computationally cheaper. This is true for a case when the kernels are computed on the fly. However, as done by \citet{cho2009kernel} and \citet{gb2018}, it is a common practice to generate the kernels using a pre-computed lookup table that maps $c^{l-1}$ to $c^l$ (in our case, map  $(c^{l-1},f)$ to $c^l$). This is done because of the large computational cost to compute a kernel when the sample size is large, even for an analytically solvable kernel.

\begin{table}[]
\caption{Performances of the sparse and dense models. "Dense - best acc.": the best-performing dense model. "Sparse - equiv. acc.": a sparse model with a performance comparable to "Dense - best acc.". "Dense - same L": a dense model with the same depth as "Sparse - equiv. acc.". The mean and standard deviation over 10 trials with randomly sampled training sets are shown.}
\centering
\begin{tabular}{l c c c c}
\toprule 
Dataset: $P$ &  & Dense - best acc. & Sparse - equiv. acc. & Dense - same $L$\tabularnewline
\cmidrule(r){1-5}
MNIST: 1000 & Accuracy & 0.9300$\pm$0.0028 & 0.9303$\pm$0.0025 & 0.9227$\pm$0.0015\tabularnewline
\cline{2-5} \cline{3-5} \cline{4-5} \cline{5-5} 
 & $L$ & 5 & 1 & 1\tabularnewline
\cline{2-5} \cline{3-5} \cline{4-5} \cline{5-5} 
 & $f$ & 0.5 & 0.139 & 0.5\tabularnewline
\hline
MNIST: 10000 & Accuracy & 0.9748$\pm$0.0014 & 0.9749$\pm$0.0013 & 0.9725$\pm$0.0013\tabularnewline
\cline{2-5} \cline{3-5} \cline{4-5} \cline{5-5} 
 & $L$ & 3 & 1 & 1\tabularnewline
\cline{2-5} \cline{3-5} \cline{4-5} \cline{5-5} 
 & $f$ & 0.5 & 0.087 & 0.5\tabularnewline
\hline
CIFAR10: 1000 & Accuracy & 0.3810$\pm$0.0033 & 0.3814$\pm$0.0046 & 0.3160$\pm$0.0060\tabularnewline
\cline{2-5} \cline{3-5} \cline{4-5} \cline{5-5} 
 & $L$ & 18 & 2 & 2\tabularnewline
\cline{2-5} \cline{3-5} \cline{4-5} \cline{5-5} 
 & $f$ & 0.5 & 0.01 & 0.5\tabularnewline
\hline 
CIFAR10: 10000 & Accuracy & 0.5016$\pm$0.0055 & 0.5017$\pm$0.0057 & 0.4621$\pm$0.0035\tabularnewline
\cline{2-5} \cline{3-5} \cline{4-5} \cline{5-5} 
 & $L$ & 18 & 3 & 3\tabularnewline
\cline{2-5} \cline{3-5} \cline{4-5} \cline{5-5} 
 & $f$ & 0.5 & 0.087 & 0.5\tabularnewline
\hline 
\end{tabular}
\label{table_result}
\end{table}

\section{Theoretical explanation of the sparsity-depth tradeoff}\label{theorySec}
\subsection{Dynamics of the kernel over layers}
As a recursive function, the kernel reaches or diverges away from a fixed point as the network gets deeper \citep{poole2016exponential,gb2017,gb2018}. Here we use the notation $q^l$ to denote the length of a representation in layer $l$ and $c^l = \cos{\theta^l}$ to denote the cosine similarity between two representations in layer $l$.

Since the activation function ReLU is unbounded, $q^l$ either decays to $0$ or explodes to $\infty$ as $l\rightarrow \infty$. However, we can find $\sigma^*$ that maintains $q^l$ at its initial length $q^1$ by setting $\sigma^{*}=\sqrt{\frac{\pi}{I\left(0\mid\tau\right)-\tau\sqrt{2\pi}}}$ which depends on the sparsity level (see Supp. \ref{app:sigma} for the full derivation). Using $\sigma^*$ is encouraged since it guarantees numerical stability in the computation of the sparse NNGP kernel, although in theory, the kernel regression is invariant to the choice of $\sigma$.

The dynamics of the $c^l$ is 
\begin{equation}
     c^{l+1}=\frac{\sigma^{*2}}{2\pi}\left(2I\left(\arccos(c^{l})\mid\tau\right)-\tau\sqrt{2\pi}(1+c^{l})\right)   
\end{equation}
when we use $\sigma^*$. This is obtained by dividing Eqn. \ref{eq:K} by $q^l=\sqrt{K^{l}(\mathbf{x}^{(p)},\mathbf{x}^{(p)})K^{l}(\mathbf{x}^{(q)},\mathbf{x}^{(q)})}$ on both sides of the equation, assuming the same norm for the inputs $p$ and $q$.

\begin{figure}
\centering
\includegraphics[width=0.9\linewidth]{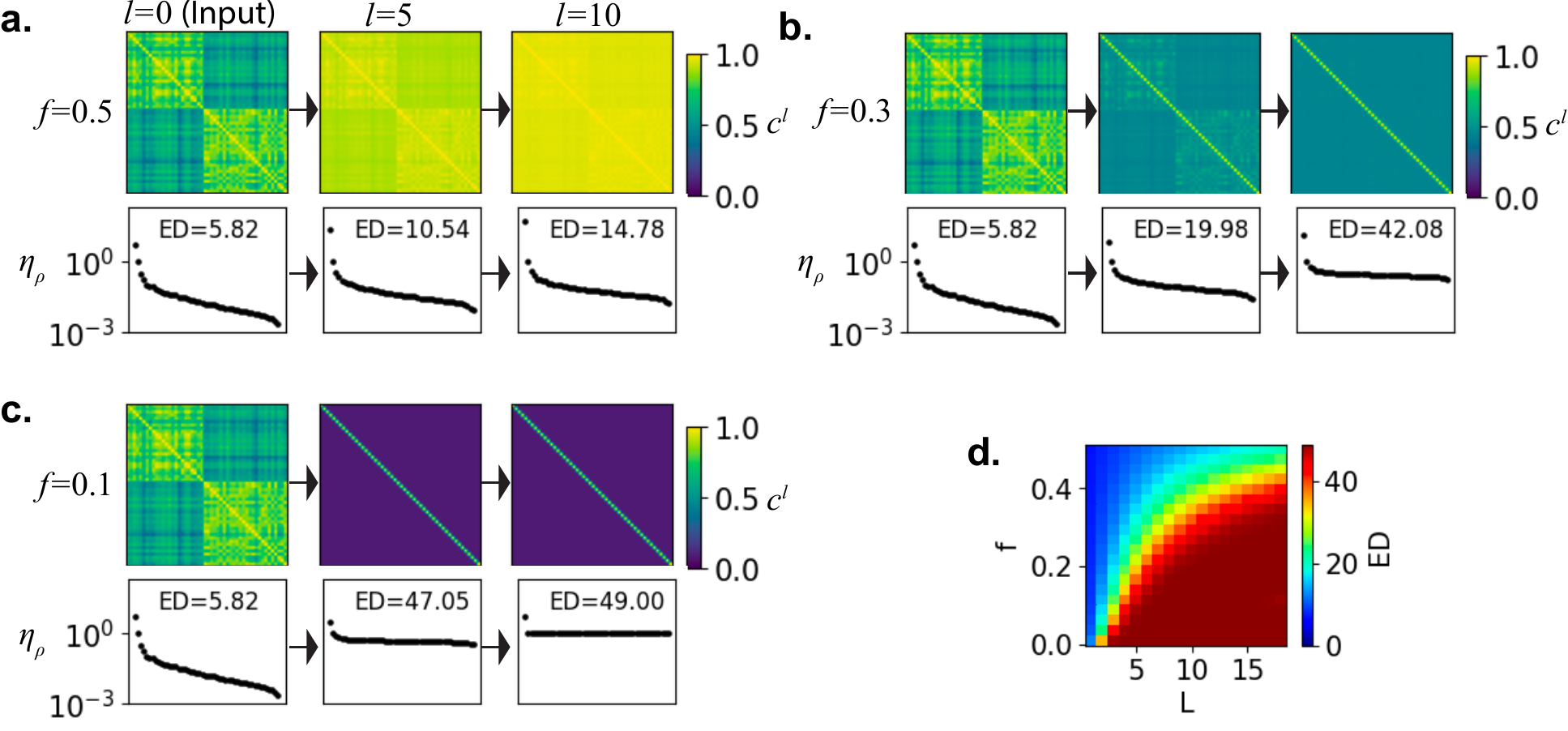}
\caption{Evolution of $\mathbb{R}^{50\times50}$ kernel Gram matrix of Fashion-MNIST data spanning two classes, i.e. shirts and ankle boots. The target function is zero-mean, since the class labels are 1 and -1. (a) $f=0.5$ case. Top: Gram matrices colored by the normalized kernel values (i.e. cosine similarities between representations $c^l$), at the input layer, $5^{th}$ layer, and $10^{th}$ layer. Bottom: the eigenspectrums of the corresponding Gram matrices normalized by the second largest eigenvalue. The effective dimensionality (ED) of a spectrum is indicated in each plot; (b) $f=0.3$ case; (c) $f=0.1$ case; (d) ED values over sparsity and depth.}
\label{dynamics}
\vspace*{-\intextsep}
\end{figure}

In Figure \ref{dynamics}, we take a Gram matrix, whose elements are $K^{l=0}(\mathbf{x}^{(p)},\mathbf{x}^{(q)})=\mathbf{x}^{(p)}\cdot \mathbf{x}^{(q)}$, spanning two categories of Fashion-MNIST dataset, and pass it through a cascade of the sparse NNGP.
We observe that the non-sparse kernel ($f=0.5$) assimilates all inputs ($c^l\rightarrow 1$) as the layer gets deeper. This means that the Gram matrix becomes a rank-1 matrix as shown in Figure \ref{dynamics}a. However, if we omit the first eigenvalue, we see that the effective dimensionality ($ED$) of the spectrum slowly increases over the layers, instead of converging to 1. We omit the first eigenvalue since it does not affect the prediction $\mu$ when the target function is zero-mean and the kernel Gram matrix has a constant function as an eigenfunction, which is commonly encountered in practice (see Supp. \ref{app:invariance}). The effective dimensionality is a participation ratio of the kernel eigenvalues $\eta_\rho$'s computed by $ED=\frac{\left(\sum_{\rho>0} \eta_\rho \right)^2}{\sum_{\rho>0}  \eta_\rho^2}$ where $\rho>0$ indicates the omission of the first eigenvalue $\eta_{0}$. This means that the eigenspectrum slowly flattens disregarding the first eigenvalue. A sparser kernel ($f=0.3$), on the other hand, decorrelates the inputs to $c^l<1$, which makes the Gram matrix become the identity matrix plus a constant non-zero off-diagonal coefficient (Figure \ref{dynamics}b). This matrix has a flat spectrum as in the case of the identity matrix, but with an offset in the first eigenvalue which reflects the non-zero off-diagonal values. Similar to the $f=0.5$ case, the $f=0.3$ case also flattens disregarding the first eigenvalue, i.e. increases $ED$, but at a faster rate. We see the flattening happens even faster for an even sparser kernel with $f=0.1$ (Figure \ref{dynamics}c). We clearly see that the flattening happens faster at sparser kernels in Figure \ref{dynamics}d that shows $ED$ over the $fL$-plane. It is noteworthy that the pattern of $ED$ over the $fL$-plane resembles that of the generalization performances shown in Figure \ref{finite} and \ref{infinite}. The next section provides a theoretical explanation that relates the shape of the kernel eigenspectrum to the generalization error.

\subsection{Generalization theory}
Here we provide a theoretical explanation of the generalization performance of the sparse NNGP kernel. We start by reviewing the theory on the in-distribution generalization error of kernel ridge regression presented by \citet{canatar2021spectral}.

\subsubsection{Background: Theory on the generalization error of kernel regression}
Assume that the target function $\bar{f}: \mathbb{R}^{n_0} \rightarrow \mathbb{R}$ exists in a reproducing kernel Hilbert space (RKHS) given by the kernel of interest. The target function can be expressed in terms of the coordinates ($\bar{v}_\rho$) on the basis of the RKHS given by the Mercer decomposition $     \int d\mathbf{x}' p(\mathbf{x}') K(\mathbf{x},\mathbf{x}') \phi_\rho (\mathbf{x}') = \eta_\rho \phi_\rho (\mathbf{x}) $
\begin{equation}
    \bar{f}(\mathbf{x})=\sum_{\rho=0}^{N-1} \bar{v}_\rho \phi_\rho(\mathbf{x}), \quad \rho = 0, \ldots, N-1.
\end{equation}
where $p(\mathbf{x}')$ is the input data distribution in $\mathbb{R}^{n_0}$, and $\phi_\rho$ and $\eta_\rho $ are the $\rho^{th}$ eigenfunction and eigenvalues respectively. We assume $N$ is infinite, which is required for the theory. At the large training sample size and large $N$ limit, the generalization error $E_g$ is expressed as a sum of modal errors $E_\rho$ weighted by the target powers $\bar{v}^2_\rho$.
\begin{equation}\label{eq:Eg}
    E_{g} =\sum_{\rho}\bar{v}_{\rho}^{2}E_{\rho}    
\end{equation}
\begin{equation}\label{eq:E_terms}
    E_{\rho} =\frac{1}{1-\gamma}\frac{\kappa^{2}}{(\kappa+P\eta_{\rho})^{2}}
    \quad\quad
    \gamma=\sum_{\rho}\frac{P\eta_{\rho}^{2}}{\left(\kappa+P\eta_{\rho}\right)^{2}}
    \quad\quad
    \kappa=\lambda+\sum_{\rho}\frac{\kappa\eta_{\rho}}{\kappa+P\eta_{\rho}}    
\end{equation}
where $P$ is the number of training samples and $\lambda$ is the ridge parameter. Note that $E_\rho$ is independent of the target function but dependent on the input distribution and kernel which are summarized in $\eta_\rho$. In Eqn.(\ref{eq:Eg}), the $E_\rho$'s are weighted by $\bar{v}^2_\rho$'s which are dependent on the target function, input distribution and kernel. Therefore in general, except for the special cases we discuss in this paper, we need to keep track of the change in both $E_\rho$ and $\bar{v}^2_\rho$, when tracking the change of $E_g$ with different kernels. Note that each $E_\rho$ is dependent on all $\eta_{\rho'}$'s whether $\rho=\rho'$ or $\rho\neq\rho'$ due to the $\kappa$ and $\gamma$ terms. $\kappa$ is a self-consistent equation that can be solved with a numerical root-finding algorithm. See Supp. \ref{app:apply} for the details of the implementation.
\begin{figure}
\centering
\includegraphics[width=1\textwidth]{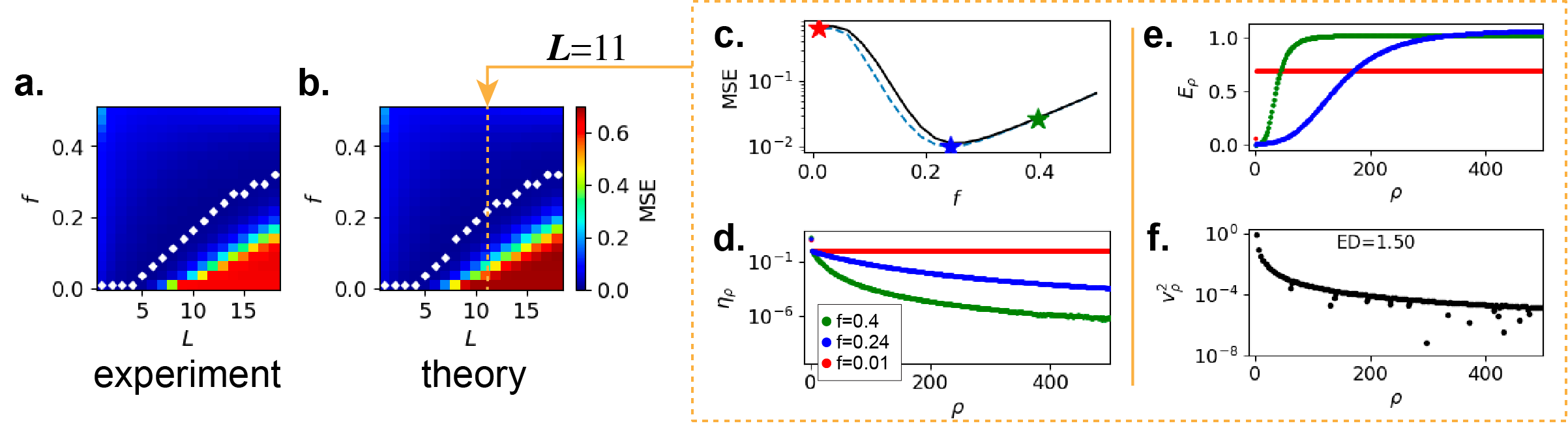}
\caption{Theoretical analysis of the generalization error over a circulant dataset of $P=1000$. (a) The Experimental result on the generalization errors over the circulant dataset over the sparsity ($f$) and depth ($L$). (b) Theoretical predictions of the generalization errors. (c) The generalization error (experimental: blue dotted line, theoretical prediction: black solid line) of the sparse kernels with depth $L=11$. The kernel with the highest, lowest, and intermediate generalization errors are indicated with red, blue, and green stars respectively; (d) The eigenspectrums (normalized by the second eigenvalues) of the kernels corresponding to the three cases marked in (c). The first 500 eigenvalues are shown; (e) The modal errors $E_\rho$  corresponding to the three cases marked in (c); (f) The target function power $\bar{v}^2_\rho$ spectrum. The effective dimensionality (ED) of the target function power spectrum is 1.5 as indicated in the figure.}
\vspace*{-\intextsep}
\label{circ}
\end{figure}

\subsubsection{Generalization theory applied to sparse NNGP}
The theoretical formulation of the generalization error accurately predicts the experimentally observed generalization errors (see Figure \ref{circ}a,b and Supp. \ref{app:real} for the real-life datasets). As an illustrating example, we use synthetic circulant data with a step target function. A Gram matrix generated from a kernel and sampled input data is a circulant matrix if the input data is evenly distributed around a circle, regardless of the choice of the kernel. The set of eigenfunctions of a circulant matrix always consists of the harmonics of the sine and cosine, so the eigenfunctions are invariant to a kernel. Therefore we need to examine the task-model alignment in terms of eigenspectrum in order to understand the generalization performance. Our target function is a zero-meaned square wave function with even step lengths.

As in the case of the real-life dataset, the circulant dataset also creates a similar generalization performance pattern over the $fL$-plane. We first inspect the spectrums of the kernels in the poor generalization performance regime (red region in Figure \ref{circ}a,b), mainly occupied by the deep sparse networks. It turns out that the eigenspectrums are flat and identical in that regime, disregarding the first eigenvalues (Figure \ref{circ}b). The first eigenvalue can vary widely in that regime, but this eigenvalue can be disregarded, since it does not affect the generalization performance (see Supp. \ref{app:invariance}). The reason we see the flat spectrums in the shallower depths for the sparser networks is that for the sparse kernels, the cosine similarity $c^l$ converges quickly to some fixed point value below $1$ as the depth becomes deeper. This means the resulting Gram matrix becomes an identity matrix offset by a value determined by the fixed point, which has a flat spectrum if we disregard the first eigenvalue that encodes the offset (Figure \ref{dynamics}).  

The kernels outside the poor performance regime have non-flat spectrums, but that does not mean the least flat spectrum performs the best. For a given depth, the best-performing kernel usually exists below $f=0.5$, which has neither the flattest nor the least flat spectrum in that given depth (Figure \ref{circ}c,d). We investigate the modal errors $E_\rho$'s of each kernel for more insight.

Note that $E_g$ is a dot product between the modal errors and the target function powers (Eqn. \ref{eq:Eg}). Also, as noted above, the target function powers do not change for the circulant dataset. Therefore, having small $E_\rho$'s for the modes that correspond to high target power $\bar{v}^2_\rho$ would greatly contribute to lowering $E_g$. On the other hand, having large $E_\rho$'s for the modes that correspond to high $\bar{v}^2_\rho$ would greatly contribute to increasing $E_g$. In practice, we also observe the increase in $E_g$ due to the increase in $E_\rho$'s that corresponds to low $\bar{v}^2_\rho$ due to a large number of such modes (see Supp. \ref{app:real} for the analysis on the real-dataset).

We observe that a spectrum with a steep drop over $\rho$ leads to modal errors with a steep increase over $\rho$ that has low $E_\rho$'s at lower $\rho$'s and high $E_\rho$'s at higher $\rho$'s (Figure \ref{circ}d,e). This relationship between the shape of the eigenspectrum and the modal error spectrum is theoretically supported in the following section. Therefore, compared to the flat eigenspectrum, a moderately steep eigenspectrum results in an optimal modal error spectrum that results in the minimum $E_g$, assuming that the target power spectrum is concentrated around the low $\rho$'s (Figure \ref{circ}e,f). However, for the same target power spectrum, if the eigenspectrum is too steep, the modal error increases too fast over $\rho$, contributing to increases in $E_g$  (Green dots in Figure \ref{circ}e).

This explanation is based on the circulant dataset with strictly invariant eigenfunctions and therefore invariant $\bar{v}_\rho^2$'s. However, we show that the same explanation can be applied to the real-life datasets, i.e. MNIST, Fashion-MNIST, CIFAR10, and CIFAR10-Gray, since the eigenfunctions for these datasets do not vary significantly over the variations in depths and sparsity of the NNGP (see Supp. \ref{app:real}).

\subsubsection{Relationship between the eigenspectrum shape and modal errors}
We expand on the generalization theory (Eqn. \ref{eq:Eg},\ref{eq:E_terms}) to elucidate how a change in the eigenspectrum affects the modal errors. To this end, we compute how $E_\rho$'s change when the spectrum is perturbed from a flat spectrum, assuming noise-free target function and $\lambda=0$. The first order perturbation of  $E_\rho$ from the flat spectrum is given by
\begin{equation} \label{eq:perturb}
    \nabla E_\rho = -2\left(1-\alpha\right)\alpha \frac{1}{\eta} \left( \nabla \eta_\rho - \langle{\nabla \eta_\rho}\rangle \right) 
\end{equation}
where $N$ is the number of non-zero eigenvalues and $\alpha=\frac{P}{N}$.  We assume $P\rightarrow \infty$ and $N\rightarrow \infty$, but $\alpha=\mathcal{O}(1)$. $\eta$ is the eigenvalue of the flat spectrum, $\nabla \eta_\rho$ the perturbation in the eigenvalue, and $\langle{\nabla \eta_\rho}\rangle = \frac{1}{N} \sum_\rho \nabla \eta_\rho$ is a mean value of the perturbations. The derivation of Eqn. \ref{eq:perturb} is presented in the Supp. \ref{app:dEg}, \ref{app:perturb}.

The intuition provided by Eqn. \ref{eq:perturb} is that the modal error perturbation $\nabla E_\rho $ is a sign-flipped version of the zero-meaned $\nabla \eta_\rho$. Therefore as the spectrum becomes less flat, the modal errors $E_\rho$'s decrease for $\rho$'s corresponding to larger eigenvalues, but $E_\rho$'s increase for $\rho$'s corresponding to smaller eigenvalues. Therefore, if the $\bar{v}_\rho$'s are band-limited to $\rho$'s corresponding to larger eigenvalues, then the generalization error $E_g$ typically decreases as the spectrum perturbs away from the flat spectrum. On the other hand, if the $\bar{v}_\rho$'s are band-limited to $\rho$'s corresponding to smaller eigenvalues, the generalization error typically increases. In practice, $\bar{v}_\rho$'s are skewed, yet spread out over the entire spectrum (See Supp. \ref{app:real}). Therefore there is a trade-off between the decrease in $E_\rho$'s at low $\rho$'s and the increase in $E_\rho$'s at high $\rho$'s. This results in requiring a moderately steep eigenspectrum, which is often quickly (over layers) achieved by sparse NNGP kernel at shallow depth (Figure \ref{dynamics}, \ref{circ}).

For input distributions where the eigenfunctions of the Gram matrix are similar between two compared kernels, the target function coefficients also exhibit similarity. In such cases, the conventional definition of task-model alignment based on eigenfunctions fails to capture the difference in $E_g$ effectively (see Supp. \ref{app:tma}). However, our approach, which considers the shape of the eigenspectrum relative to the target function coefficients, successfully captures this difference. This finding complements the spectral bias result presented by \citet{canatar2021spectral}. The results on spectral bias by \citet{canatar2021spectral} show that the $E_\rho$'s are in ascending order in contrast to the descending order of $\eta_\rho$, and $E_\rho$ that corresponds to greater $\eta_\rho$ decay at the faster rate as $P$ increases. However, we need more than the fact that $E_\rho$ monotonically increases over $\rho$, to compare $E_\rho$ spectrum between kernels since it does not offer an absolute reference for the comparison. Our analysis offers a way to compare spectral biases by computing the explicit first-order perturbation in $E_\rho$'s, allowing a comparison between kernels theoretically more tractable and interpretable.

\section{Discussion}

We have demonstrated that random sparse representation in a wide neural network enhances generalization performance in the NNGP limit at shallow depths. Our sparse NNGP achieves comparable performance to deep and dense NNGP (i.e., arccosine kernel) with reduced kernel composition. The performance of the arccosine kernel is known to be comparable to finite neural networks trained with stochastic gradient descent in certain contexts \citep{gb2018,lee2020finite}.

Our analysis reveals that the kernel Gram matrices for sparse and shallow (and dense and deep) networks have an eigenspectrum that yields a model error spectrum that is well-aligned with typical target functions. As demonstrated in the main section, our extended theory on the generalization of kernel regression facilitates the comparison of generalization performance between any two kernels.

Using sparse and shallow architecture, \cite{babadi2014sparseness} shows improvements in classification tasks enabled by sparsity in the cerebellum. Our results indicate that sparsity also improves regression performance, which could benefit the computation in the cortex and cerebellum.

Further investigations are needed to explore sparsity's impact in networks with learned representations. Developing a finite-width correction for our sparse NNGP kernel would enable examining sparsity effects within trained representations. Additionally, the implications of sparsity in the trainable intermediate infinite-width layers of the Neural Tangent Kernel should be considered. Furthermore, exploring sparsity's influence in different neural network architectures, including convolutional neural networks, has potential for future research.

\section*{Acknowledgements}

\bibliographystyle{unsrtnat}
\bibliography{biblio}

\newpage
\appendix
\renewcommand{\thefigure}{S\arabic{figure}}
\renewcommand{\thetable}{S\arabic{table}}
\setcounter{figure}{0}    
\setcounter{table}{0}    

\section{What Does Sparsity Mean Conceptually in the Infinitely Wide Neural Network?}\label{app:concept}

An intuitive definition of an NNGP kernel is: the dot product of $n$ dimensional neuronal representations of input $x^{(1)}$ and $x^{(2)}$, averaged over all possible realizations of a random neural network. For each realization, we take the dot product between neural representations of two inputs.

For the sparse NNGP, one might mistakenly speculate that the same subset of neurons is active for all inputs (in each realization of the network). For inputs $x^{(1)}$ and $x^{(2)}$, different sets of neurons are non-zero, since we are only constraining the fraction of active neurons, not which neurons are active.

If we constrain which subset of neurons are always active and the rest inactive, then $n=100$ with $f=0.3$ is equivalent to a narrow network of $n=30$ with $f=1$ which is essentially a linear network. In this pathological scenario, the choice of $f$ does not matter in the $n\rightarrow \infty$ limit as the reviewer surmised. Generally, however, a ReLU neural network activates different sets of neurons for different inputs, which gives it an interesting nonlinear property.

\section{Sparse NNGP Kernel Derivation}\label{app:NNGP_deriv}

The similarity of two neural representation $p$ and $q$ is $K^{l}(\mathbf{x}^{(p)},\mathbf{x}^{(q)})$
which is evaluated with the following integration.

\begin{equation}
K^{l}(\mathbf{x}^{(p)},\mathbf{x}^{(q)})=\int d\mathbf{w}_{j}P(\mathbf{w}_{j})\left[\mathbf{w}_{j}\cdot \mathbf{x}^{(p),l-1}-b\right]_+\left[\mathbf{w}_{j}\cdot \mathbf{x}^{(q),l-1}-b\right]_+
\end{equation}

At $N\rightarrow\infty$, we can invoke the central limit theorem and express the integration
in terms of a normal random variable $h_{j}^{(p)}=\mathbf{w}_{j}\cdot\mathbf{x}^{(p),l-1}$.
The covariance of the preactivation between two stimuli ($\text{Cov}[\mathbf{w}\cdot x^{(p),l-1},\mathbf{w}\cdot x^{(q),l-1}]$)
is denoted $\sigma_{pq}^{2}$ and the variance $\sigma_{p}^{2}$.

The above integration becomes

\begin{equation}
K^{l}(\mathbf{x}^{(p)},\mathbf{x}^{(q)})=\frac{1}{Z}\int d\mathbf{a}\exp\text{\ensuremath{\left(-\frac{1}{2}\begin{bmatrix}a^{(p)}  &  a^{(q)}\end{bmatrix}\Sigma^{-1}\begin{bmatrix}a^{(p)}  &  a^{(q)}\end{bmatrix}^{T}\right)}}\left[a^{(p)}-b\right]_{+}\left[a^{(q)}-b\right]_{+}
\end{equation}

\begin{equation}
\Sigma=\begin{bmatrix}\sigma_{p}^{2} & \sigma_{pq}^{2}\\
\sigma_{pq}^{2} & \sigma_{q}^{2}
\end{bmatrix}
\end{equation}
The covariance makes it difficult to solve the integration, so we
change the basis of $\mathbf{a}=\begin{bmatrix}a^{(p)} & a^{(q)}\end{bmatrix}^{\top}$
such that the we can instead integrate over a pair of independent
normal random variables $z_{1}$ and $z_{2}$. Solve this system
of equations.
\begin{equation}
a^{(p)}=\mathbf{s}_{p}\cdot\mathbf{z}
\end{equation}

\begin{equation}
a^{(q)}=\mathbf{s}_{q}\cdot\mathbf{z}
\end{equation}
The vectors $\mathbf{s}_{p}$ and $\mathbf{s}_{q}$ are expressed
in terms of the variances and covariance.

\begin{equation}
\mathbf{s}_{p}=\begin{bmatrix}\frac{\sqrt{\sigma_{p}^{2}\sigma_{q}^{2}-\sigma_{pq}^{4}}}{\sigma_{q}} & \frac{\sigma_{pq}^{2}}{\sigma_{q}}\end{bmatrix}
\end{equation}

\begin{equation}
\mathbf{s}_{q}=\begin{bmatrix}0 & \sigma_{q}\end{bmatrix}
\end{equation}
The norms of these vectors are $\|\mathbf{s}_{p}\|=\sigma_{p}$
and $\|\mathbf{s}_{q}\|=\sigma_{q}$. 

\begin{equation}
K^{l}(\mathbf{x}^{(p)},\mathbf{x}^{(q)})=\frac{1}{2\pi}\int d\mathbf{z}\exp\text{\ensuremath{\left(-\frac{1}{2}\mathbf{z}^{T}\mathbf{z}\right)}}\left[\mathbf{s}_{p}\cdot\mathbf{z}-b\right]_{+}\left[\mathbf{s}_{q}\cdot\mathbf{z}-b\right]_{+}\label{eq:before_rotation}
\end{equation}
From the previous section, we know $b_{p}=\sigma_{p}\tau$. As noted
above, $\|\mathbf{s}_{p}\|=\sigma_{p}$, so $b_{p}=\|\mathbf{s}_{p}\|\tau$.

Now perform the Gaussian integral. Unfortunately, there is no closed-form
solution, However, inspired by the derivation of the sparse step-function NNGP presented in \citetSM{cho2011analysis}, we can express this as a 1D integral which is significantly
simpler to numerically calculate. Start by adopting a new coordinate
system with basis $\mathbf{e}_{1}$ and $\mathbf{e}_{2}$, where
$\mathbf{e}_{1}$ is aligned with $\mathbf{s}_{p}$. In that
coordinate system $\mathbf{s}_{p}=\begin{bmatrix}\|\mathbf{s}_{p}\| & 0\end{bmatrix}$,
and therefore $\mathbf{s}_{p}\cdot\mathbf{z}=z_{1}\|\mathbf{s}_{p}\|$.
For $\mathbf{s}_{q}$ we have $\mathbf{s}_{q}=\begin{bmatrix}\|\mathbf{s}_{q}\|\cos\theta & \|\mathbf{s}_{q}\|\sin\theta\end{bmatrix}$,
therefore $\mathbf{s}_{q}\cdot\mathbf{z}=z_{1}\|\mathbf{s}_{q}\|\cos\theta+z_{2}\|\mathbf{s}_{q}\|\sin\theta$.
With these substitutions, Eqn. \ref{eq:before_rotation} becomes the
following.

\begin{align}
K^{l}(\mathbf{x}^{(p)},\mathbf{x}^{(q)})=&\frac{1}{2\pi}\int\int dz_{1}dz_{2}\exp\text{\ensuremath{\left(-\frac{1}{2}z_{1}^{2}+z_{2}^{2}\right)}}\left[z_{1}\|\mathbf{s}_{p}\|-b\right]_{+}\left[z_{1}\|\mathbf{s}_{q}\|\cos\theta+z_{2}\|\mathbf{s}_{q}\|\sin\theta-b\right]_{+}\label{eq:subs1}
\\
=&\frac{1}{2\pi}\int\int dz_{1}dz_{2}\exp\text{\ensuremath{\left(-\frac{1}{2}z_{1}^{2}+z_{2}^{2}\right)}}\left[z_{1}\|\mathbf{s}_{p}\|-\|\mathbf{s}_{p}\|\tau\right]_{+}\left[z_{1}\|\mathbf{s}_{q}\|\cos\theta+z_{2}\|\mathbf{s}_{q}\|\sin\theta-\|\mathbf{s}_{q}\|\tau\right]_{+}
\\
=& \frac{\|\mathbf{s}_{p}\|\|\mathbf{s}_{q}\|}{2\pi}\int\int dz_{1}dz_{2}\exp\text{\ensuremath{\left(-\frac{1}{2}z_{1}^{2}+z_{2}^{2}\right)}}\left[z_{1}-\tau\right]_{+}\left[z_{1}\cos\theta+z_{2}\sin\theta-\tau\right]_{+}    
\end{align}

where

\begin{equation}
\frac{\|\mathbf{s}_{p}\|\|\mathbf{s}_{q}\|}{2\pi}=\frac{1}{2\pi}\sigma_{p}\sigma_{q}
\end{equation}

Adopt the polar coordinate system. $z_{1}^{2}+z_{2}^{2}=r^{2}$, $r\cos\phi=z_{1}$,
$r\sin\phi=z_{2}$, $rdrd\phi=dz_{1}dz_{2}$.

\begin{align}
K^{l}(\mathbf{x}^{(p)},\mathbf{x}^{(q)})=&\frac{1}{2\pi}\sigma_{p}\sigma_{q}\int_{-\pi}^{\pi}d\phi\int_{0}^{\infty}dr\exp\text{\ensuremath{\left(-\frac{r^{2}}{2}\right)}}r\left[r\cos\phi-\tau\right]_{+}\left[r\cos\phi\cos\theta+r\sin\phi\sin\theta-\tau\right]_{+}
\\
=&\frac{1}{2\pi}\sigma_{p}\sigma_{q}\int_{-\pi}^{\pi}d\phi\int_{0}^{\infty}dr\exp\text{\ensuremath{\left(-\frac{r^{2}}{2}\right)}}r\left[r\cos\phi-\tau\right]_{+}\left[r\cos(\phi-\theta)-\tau\right]_{+}
\end{align}

Solve the integration of $\int dr\exp\text{\ensuremath{\left(-\frac{r^{2}}{2}\right)}}r\left[r\cos\phi-\tau\right]_{+}\left[r\cos(\phi-\theta)-\tau\right]_{+}$
within the range of $r$ where the integrand is non-zero. Since we
have not yet found that range, here we just perform an indefinite
integral. We will find the range and apply it later.

\begin{align}
&\int dr\exp\left(-\frac{r^{2}}{2}\right)r(r\cos\phi-\tau)(r\cos(\phi-\theta)-\tau)
\\
=&-\frac{\sqrt{2\pi}}{2}\tau(\cos\phi+\cos(\phi-\theta))\text{erf}\left(\frac{r}{\sqrt{2}}\right)-e^{-\frac{r^{2}}{2}}\left(\cos\phi\cos(\phi-\theta)\left(r^{2}+2\right)-\cos\phi r\tau+\tau(\tau-\cos(\phi-\theta)r)\right)
\\
=&-\frac{\sqrt{2\pi}}{2}\tau(\cos\phi+\cos(\phi-\theta))\text{erf}\left(\frac{r}{\sqrt{2}}\right)
\\
&\quad-e^{-\frac{r^{2}}{2}}\left(\cos\phi\cos(\phi-\theta)r^{2}+2\cos\phi\cos(\phi-\theta)-\tau\cos\phi r+\tau^{2}-\tau\cos(\phi-\theta)r\right)
\\
=&-\frac{\sqrt{2\pi}}{2}\tau(\cos\phi+\cos(\phi-\theta))\text{erf}\left(\frac{r}{\sqrt{2}}\right)-e^{-\frac{r^{2}}{2}}\left((\cos\phi r-\tau)(\cos(\phi-\theta)r-\tau)+2\cos\phi\cos(\phi-\theta)\right)
\\
=&-\exp-\frac{r^{2}}{2}\left(\left(r\cos\phi-\tau\right)\left(r\cos(\theta-\phi)-\tau\right)+2\cos\phi\cos(\theta-\phi)\right)-\tau\left(\cos\phi+\cos(\theta-\phi)\right)\sqrt{\frac{\pi}{2}}\text{erf\ensuremath{\left(\frac{r}{\sqrt{2}}\right)}}\label{eq:indef}
\end{align}
The feasible range is $r\cos\phi-\tau>0$ and $r\cos(\phi-\theta)-\tau>0$.
Assume $\tau>0$. That means at least $\cos\phi>0$ and $\cos(\phi-\theta)>0$
for all cases (but this is not a sufficiant condition).

\begin{equation}
r>\frac{\tau}{\cos\phi}
\end{equation}

\begin{equation}
r>\frac{\tau}{\cos(\phi-\theta)}
\end{equation}
Therefore

\begin{equation}
r>\max\left(\frac{\tau}{\cos\phi},\frac{\tau}{\cos(\phi-\theta)}\right)
\end{equation}
Find at what $\phi$ the inequality $\frac{\tau}{\cos\phi}<\frac{\tau}{\cos(\phi-\theta)}$
holds. Note that $-\frac{\pi}{2}<\phi<\frac{\pi}{2}$ (from $\cos\phi>0$)
and $0<\theta<\pi$.

\begin{equation}
\cos(\phi-\theta)<\cos\phi
\end{equation}
which is equivalent to

\begin{equation}
\phi<\frac{\theta}{2}=\phi_{c}\label{eq:phi_c}
\end{equation}
Find the range of $\phi$. From $\cos\phi>0$ we have $-\frac{\pi}{2}\leq\phi<\frac{\pi}{2}$.
From $\cos\left(\theta-\phi\right)>0$ we have $-\frac{\pi}{2}\leq\phi-\theta<\frac{\pi}{2}$,
which is $\theta-\frac{\pi}{2}\leq\phi<\theta+\frac{\pi}{2}$. The
intersecting domain of the two inequalities is:

\begin{equation}
\theta-\frac{\pi}{2}\leq\phi<\frac{\pi}{2}
\end{equation}
Now apply these ranges to Eqn. \ref{eq:indef}. At $r=\infty$, the
indefinite integral is:

\begin{equation}
-\tau\left(\cos\phi+\cos(\theta-\phi)\right)\sqrt{\frac{\pi}{2}}
\end{equation}
which is for the range $-\frac{\pi}{2}\leq\phi<\frac{\pi}{2}$.

For the range $\theta-\frac{\pi}{2}\leq\phi<\phi_{c}$, $r$ is integrated
from $\frac{\tau}{\cos(\phi-\theta)}$ to $\infty$. For the range
$\phi_{c}\leq\phi<\frac{\pi}{2}$, $r$ is integrated from $\frac{\tau}{\cos\phi}$
to $\infty$.

When $r=\frac{\tau}{\cos(\phi-\theta)}$, the indefinite integral
(Eqn. \ref{eq:indef}) is:

\begin{equation}
L_{1}=-\exp\left(-\frac{\tau^{2}}{2\cos^{2}(\phi-\theta)}\right)2\cos\phi\cos(\theta-\phi)-\tau\left(\cos\phi+\cos(\theta-\phi)\right)\sqrt{\frac{\pi}{2}}\text{erf\ensuremath{\left(\frac{\tau}{\sqrt{2}\cos(\phi-\theta)}\right)}}
\end{equation}
When $r=\frac{\tau}{\cos\phi}$, the indefinite integral (Eqn. \ref{eq:indef})
is:

\begin{equation}
L_{2}=-\exp\left(-\frac{\tau^{2}}{2\cos^{2}\phi}\right)2\cos\phi\cos(\theta-\phi)-\tau\left(\cos\phi+\cos(\theta-\phi)\right)\sqrt{\frac{\pi}{2}}\text{erf\ensuremath{\left(\frac{\tau}{\sqrt{2}\cos\phi}\right)}}
\end{equation}
Therefore the definite integral is:

\begin{equation}
K^{l}(\mathbf{x}^{(p)},\mathbf{x}^{(q)})=\frac{1}{2\pi}\sigma_{p}\sigma_{q}\left(\int_{\theta-\frac{\pi}{2}}^{\pi/2}-\tau\left(\cos\phi+\cos(\theta-\phi)\right)d\phi\sqrt{\frac{\pi}{2}}-\left(\int_{\theta-\frac{\pi}{2}}^{\phi_{c}}L_{1}d\phi+\int_{\phi_{c}}^{\pi/2}L_{2}d\phi\right)\right)\label{eq:def_int}
\end{equation}
In the following steps, we clean up the expression $\int_{\theta-\frac{\pi}{2}}^{\phi_{c}}L_{1}d\phi$.

\begin{equation}
I_{1}=-\int_{\theta-\frac{\pi}{2}}^{\phi_{c}}L_{1}d\phi
\end{equation}
Substitue $\phi$ with $\phi=\phi_{0}+\theta-\frac{\pi}{2}$.

\begin{equation}
I_{1}=\int_{0}^{\phi_{c}-\theta+\frac{\pi}{2}}\exp\left(-\frac{\tau^{2}}{2\sin^{2}(\phi_{0})}\right)2\sin\left(\phi_{0}+\theta\right)\sin(\phi_{0})+\tau\left(\sin\left(\phi_{0}+\theta\right)+\sin(\phi_{0})\right)\sqrt{\frac{\pi}{2}}\text{erf\ensuremath{\left(\frac{\tau}{\sqrt{2}\sin(\phi_{0})}\right)}\ensuremath{d\ensuremath{\phi_{0}}}}
\end{equation}
Also clean up the expression for $\int_{\phi_{c}}^{\pi/2}L_{2}d\phi$.

\begin{equation}
I_{2}=-\int_{\phi_{c}}^{\pi/2}L_{2}d\phi=\int_{\pi/2}^{\phi_{c}}L_{2}d\phi
\end{equation}
Substitue $\phi$ with $\phi=\phi_{0}+\frac{\pi}{2}$

\begin{equation}
I_{2}=\int_{0}^{\frac{\pi}{2}-\phi_{c}}\exp\left(-\frac{\tau^{2}}{2\sin^{2}(\phi_{0})}\right)2\sin(\phi_{0})\sin(\theta+\phi_{0})+\tau\left(\sin(\phi_{0})+\sin(\theta+\phi_{0})\right)\sqrt{\frac{\pi}{2}}\text{erf\ensuremath{\left(\frac{\tau}{\sqrt{2}\sin(\phi_{0})}\right)}}d\phi_{0}
\end{equation}
Notice that the integrand for $I_{1}$ and $I_{2}$ are the same,
and only the upper bonds of the integration ranges are different.

Solve $\int_{\theta-\frac{\pi}{2}}^{\pi/2}-\tau\left(\cos\phi+\cos(\theta-\phi)\right)d\phi\sqrt{\frac{\pi}{2}}$
term in Eqn. \ref{eq:def_int}.

\begin{align}
\sqrt{\frac{\pi}{2}}\int_{\theta-\frac{\pi}{2}}^{\pi/2}-\tau\left(\cos\phi+\cos(\theta-\phi)\right)d\phi
\\
=&-\tau\sqrt{\frac{\pi}{2}}\left(\sin\phi+\sin(\phi-\theta)\right)\bigg|_{\theta-\frac{\pi}{2}}^{\pi/2}
\\
=&-\tau\sqrt{\frac{\pi}{2}}\left(\left(1+\cos(\theta)\right)-\left(-\cos(\theta)-1\right)\right)
\\
=&-\tau\sqrt{2\pi}(1+\cos\theta)
\end{align}
Therefore, Eqn. \ref{eq:def_int} is equivalent to

\begin{equation}
K^{l}(\mathbf{x}^{(p)},\mathbf{x}^{(q)})=\frac{1}{2\pi}\sigma_{p}\sigma_{q}\left(I_{1}+I_{2}-\tau\sqrt{2\pi}(1+\cos\theta)\right)
\end{equation}
As shown in Eqn. \ref{eq:phi_c}, $\frac{\theta}{2}=\phi_{c}$. Substite
$\phi_{c}$ with $\frac{\theta}{2}$ in $I_{1}$ and $I_{2}$. Therefore
the final expression for the similarity is

\begin{equation}
K^{l}(\mathbf{x}^{(p)},\mathbf{x}^{(q)})=\frac{1}{2\pi}\sigma_{p}\sigma_{q}\left(2I(\theta\mid\tau)-\tau\sqrt{2\pi}(1+\cos\theta)\right)\label{eq:simil1}
\end{equation}
where
\begin{equation}
I(\theta\mid\tau)=\int_{0}^{\frac{\pi-\theta}{2}}\exp\left(-\frac{\tau^{2}}{2\sin^{2}(\phi_{0})}\right)2\sin\left(\phi_{0}+\theta\right)\sin(\phi_{0})+\tau\left(\sin\left(\phi_{0}+\theta\right)+\sin(\phi_{0})\right)\sqrt{\frac{\pi}{2}}\text{erf\ensuremath{\left(\frac{\tau}{\sqrt{2}\sin(\phi_{0})}\right)}\ensuremath{d\ensuremath{\phi_{0}}}}\label{eq:int}
\end{equation}
The intergration term (Eqn. \ref{eq:int}) can be efficiently computed
using a simple numerical integration.

In the step where we make a substitution in Eqn. \ref{eq:subs1},
the angle $\theta$ is defined as an angle between $\mathbf{s}_{p}$
and $\mathbf{s}_{q}$ which is an angle between neural representations.

\begin{equation}
\theta=\arccos\frac{\mathbf{s}_{p}\cdot\mathbf{s}_{q}}{\|\mathbf{s}_{p}\|\|\mathbf{s}_{q}\|}=\arccos\frac{\sigma_{pq}^{2}}{\sigma_{p}\sigma_{q}}\label{eq:theta_arcos}
\end{equation}
For the kernel of the first hidden layer, $\sigma_{p}^{2}=\frac{\sigma^{2}}{N}\|\mathbf{x}^{(p)}\|^{2}$, and $\sigma_{pq}^{2}=\frac{\sigma^{2}}{N}\|\mathbf{x}^{(p)}\|\|\mathbf{x}^{(q)}\|$. Therefore, the above equation (Eqn. \ref{eq:theta_arcos}) becomes

\begin{equation}
\theta=\arccos\left[\frac{\mathbf{x}^{(p)}\cdot\mathbf{x}^{(q)}}{\|\mathbf{x}^{(p)}\|\|\mathbf{x}^{(q)}\|}\right]\label{eq:theta1}
\end{equation}

For the deeper layers, we have $\sigma_{p}^{2}=\sigma^2 K^{l-1}(\mathbf{x}^{(p)},\mathbf{x}^{(p)})$, and $\sigma_{pq}^{2}=\sigma^2 K^{l-1}(\mathbf{x}^{(p)},\mathbf{x}^{(q)})$. With this substitution, we arrive at the general solution presented in the main section of the paper.

\begin{equation} 
    K^{l}(\mathbf{x}^{(p)},\mathbf{x}^{(q)}) =\frac{\sigma^{2}}{2\pi}\sqrt{K^{l-1}(\mathbf{x}^{(p)},\mathbf{x}^{(p)})K^{l-1}(\mathbf{x}^{(q)},\mathbf{x}^{(q)})} \left(2I\left(\theta^{l}\mid\tau\right)- \tau\sqrt{2\pi}(1+\cos\theta^{l})\right)
\end{equation}
\begin{equation}
    \theta^{l}=\arccos\frac{K^{l-1}(\mathbf{x}^{(p)},\mathbf{x}^{(q)})}{\sqrt{K^{l-1}(\mathbf{x}^{(p)},\mathbf{x}^{(p)})K^{l-1}(\mathbf{x}^{(q)},\mathbf{x}^{(q)})}}
\end{equation}

\section{$\sigma^{*}$ Derivation}\label{app:sigma}

The sparse kernel equation is shown below.

\begin{equation}
K^{l+1}(\mathbf{x}^{(p)},\mathbf{x}^{(q)})=\frac{\sigma^{2}}{2\pi}\sqrt{K^{l}(\mathbf{x}^{(p)},\mathbf{x}^{(p)})K^{l}(\mathbf{x}^{(q)},\mathbf{x}^{(q)})}\left(2I\left(\theta^{l}\mid\tau\right)-\tau\sqrt{2\pi}(1+\cos\theta^{l})\right)
\end{equation}

\begin{equation}
\theta^{l}=\arccos\frac{K^{l}(\mathbf{x}^{(p)},\mathbf{x}^{(q)})}{\sqrt{K^{l}(\mathbf{x}^{(p)},\mathbf{x}^{(p)})K^{l}(\mathbf{x}^{(q)},\mathbf{x}^{(q)})}}
\end{equation}
Since we want to see the evolution of the representation length $K^{l}(\mathbf{x}^{(p)},\mathbf{x}^{(p)})$,
substitute $\mathbf{x}^{(q)}$ with $\mathbf{x}^{(p)}$ in the
kernel equation. We should use $\theta=0$ at all layers.

\begin{equation}
K^{l+1}(\mathbf{x}^{(p)},\mathbf{x}^{(p)})=\frac{\sigma^{2}}{2\pi}K_{h}^{l}(\mathbf{x}^{(p)},\mathbf{x}^{(p)})\left(2I\left(0\mid\tau\right)-\tau2\sqrt{2\pi}\right)
\end{equation}
We require that the representation
length do not change over layers, i.e. $K_{h}^{l}(\mathbf{x}^{(p)},\mathbf{x}^{(p)})=K_{h}^{l+1}(\mathbf{x}^{(p)},\mathbf{x}^{(p)})$.
Therefore,

\begin{equation}
K_{h}^{l}(\mathbf{x}^{(p)},\mathbf{x}^{(p)})=\frac{\sigma^{*2}}{\pi}K_{h}^{l}(\mathbf{x}^{(p)},\mathbf{x}^{(p)})\left(I\left(0\mid\tau\right)-\tau\sqrt{2\pi}\right)
\end{equation}

\begin{equation}
\left[1-\frac{\sigma^{*2}}{2\pi}\left(2I\left(0\mid\tau\right)-\tau2\sqrt{2\pi}\right)\right]K_{h}^{l}(\mathbf{x}^{(p)},\mathbf{x}^{(p)})=0
\end{equation}
This means the following equality must be satisfied.

\begin{equation}
1=\frac{\sigma^{*2}}{\pi}\left(I\left(0\mid\tau\right)-\tau\sqrt{2\pi}\right)
\end{equation}
This equality can always be satisfied, by computing $\sigma^{*}$
given $\tau$.

\begin{equation}
\sigma^{*}=\sqrt{\frac{\pi}{I\left(0\mid\tau\right)-\tau\sqrt{2\pi}}}
\end{equation}

\section{Relationship to the Arcosine Kernel}\label{app:relate}

$\tau$ in Eq. (5) and (8) is the variable that is dependent on the sparsity $f$ ($\tau = \sqrt{2} erf ^{-1}(1-2f)$, where $erf^{-1}$ is the inverse error function). This means $\tau=0$ when $f=0.5$, and $\tau\rightarrow\infty$ as $f\rightarrow 0$. The arccosine kernel presented in the previous works by \citetSM{cho2009kernel} and \citetSM{gb2018} is the case when $\tau=0$ ($f=0.5$).

We show that the integration term Eqn. \ref{eq:I} reduces to a simpler form when $\tau=0$.

$$I(\theta\mid0)=\int_{0}^{\frac{\pi-\theta}{2}}2\sin\left(\phi_{0}+\theta\right)\sin(\phi_{0})\ensuremath{d\phi_{0}}$$

$$=\frac{1}{2}\left(2\phi_{0}\cos(\theta)-\sin\left(\theta+2\phi_{0}\right)\right)\vert_{0}^{\frac{\pi-\theta}{2}}$$

$$=\frac{1}{2}\left(\left(\pi-\theta\right)\cos(\theta)+\sin\left(\theta\right)\right)$$
Therefore, Eqn. \ref{eq:K0} simplifies to
$$ K(\mathbf{x}^{(p)},\mathbf{x}^{(q)}) \\ =\frac{\sigma^{2}}{2\pi}\|\mathbf{x}^{(p)}\|\|\mathbf{x}^{(q)}\| \left(\left(\pi-\theta\right)\cos(\theta)+\sin\left(\theta\right)\right)$$
which matches Eqn. (3), (6) in \citetSM{cho2009kernel} up to a scaling factor $\frac{\sigma^2}{2}$ which is an arbitrary choice.
 
\subsection{Sensitivity of the Arcosine Kernel to the Bias Noise}\label{app:bnoise}

\begin{figure}
\centering
\includegraphics[width=0.7\linewidth]{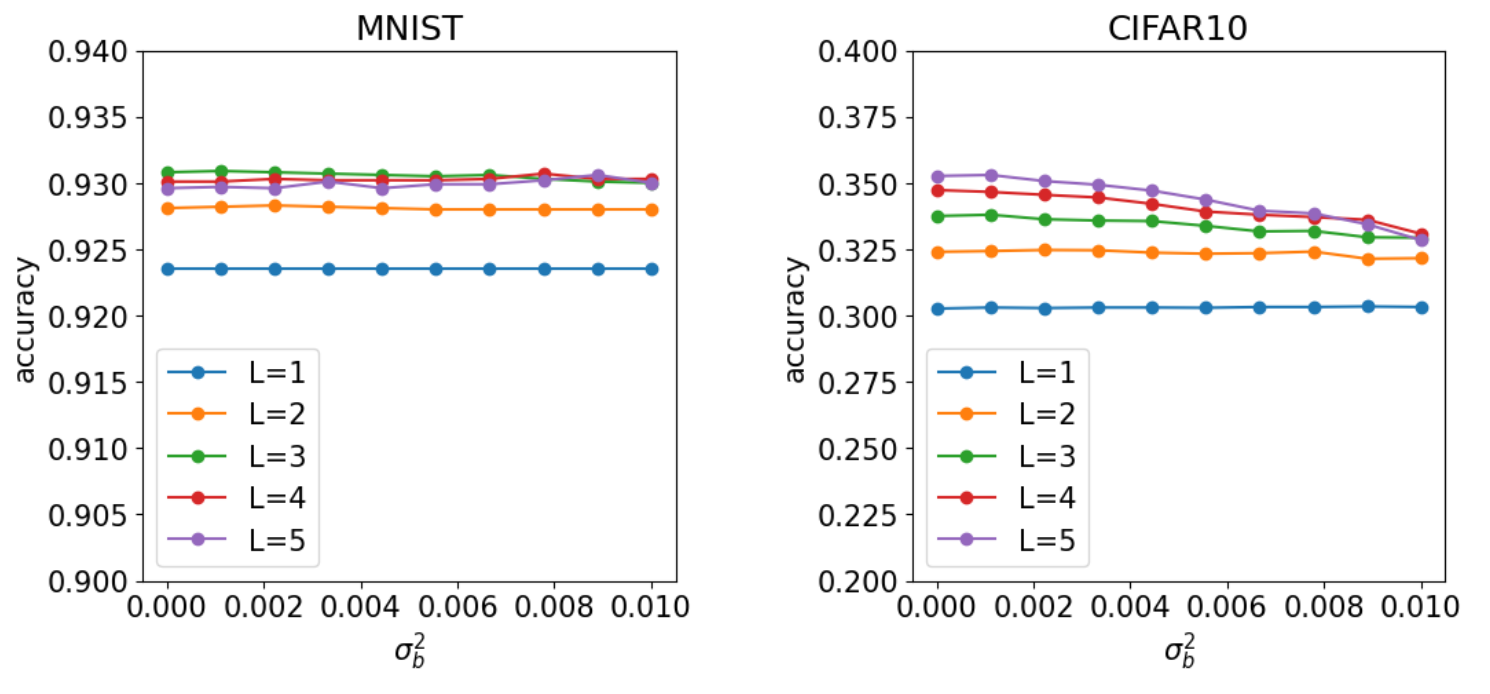}
\caption{The generalization performance (accuracy) is not significantly affected by the magnitude of the bias noise. Moreover, the larger magnitude degrades the generalization performance in CIFAR10.}
\label{fig:bnoise}
\vspace*{-\intextsep}
\end{figure}

The effect of tuning the standard deviation of the bias $\sigma_b^2$ in the arccosine kernel is negligible, as provided in \citetSM{gb2018}. Here we provide the empirical and theoretical evidence for certain setups (Figure \ref{fig:bnoise}). In the $f=0.5$ case, it has been shown empirically in Figure 4b (and supplementary Figure 9) of \citetSM{gb2018}, that $\sigma_b^2$ does not significantly affect the generalization performance. Discussing the case when $f<0.5$ is irrelevant here, since we need a constant bias (as opposed to a random bias) in order to keep the desired sparsity level.

We can show this theoretically for a single-hidden layer case. As shown in Eqn. 5 of Lee et al., 2018, a non-zero $\sigma^2_b$ effectively offsets the value of the kernel by exactly $+\sigma^2_b$. We show in the main text of our manuscript that in a usual setting, the offset of the kernel does not affect the generalization performance. Hence we can theoretically show in the single hidden layer case, the choice of $\sigma^2_b$ does not usually affect the generalization performance.

\section{Effect of the Regularization on the Performance of the Sparse NNGP}\label{app:regular}

\begin{figure}
\centering
\includegraphics[width=0.5\linewidth]{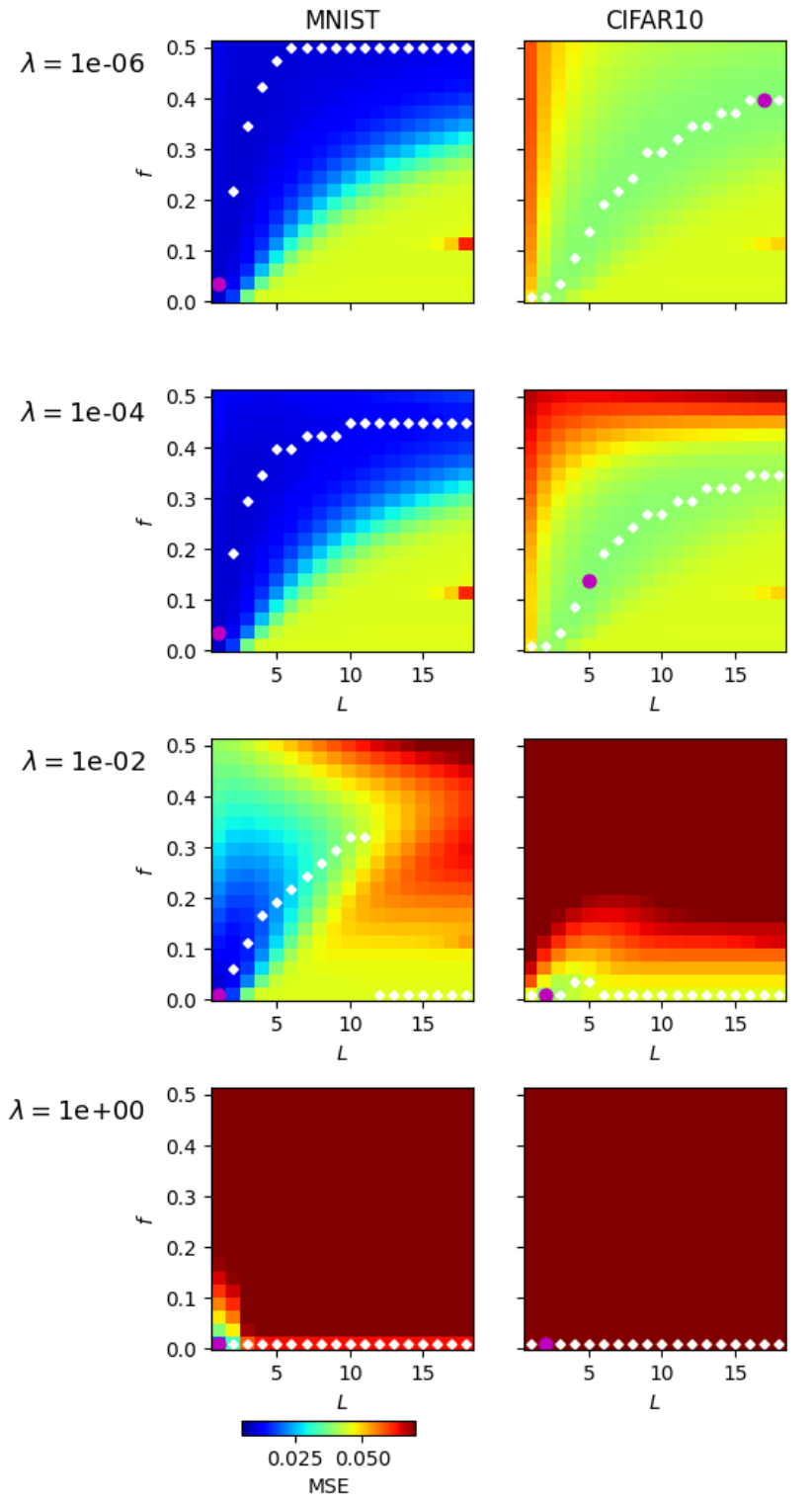}
\caption{The generalization performance (MSE) for different choices of regularization factor of the kernel ridge-regression (each row). As shown above, the larger ridge parameter makes the sparser NNGP kernels perform even better than the non-sparse counterpart. This observation is pronounced and consistent.}
\label{eff_lamb}
\vspace*{-\intextsep}
\end{figure}

The observation noise, i.e. ridge $\lambda$, is not a part of the model parameter and requires a dedicated investigation, which is a possible future direction. Here, we share an example figure that shows the effect of $\lambda$ on the generalization performance (Figure \ref{eff_lamb}). It is evident that larger the $\lambda$ makes sparser kernels perform even better than the non-sparse counterparts. This observation is pronounced and consistent.

\section{Details of the numerical experiments}\label{app:exp}
For the main results, MNIST, Fashion-MNIST, CIFAR10, and a grayscale version of CIFAR10 are used. From each dataset, $P$ number of samples are randomly chosen as training samples. The images are flattened before being used as inputs. We use one-hot vector $\mathbb{R}^{10}$ representations of the class labels. The random sampling of the training data is repeated to check the consistency of our results.

The training is done using kernel ridge regression. We generate the kernels for training and test datasets and use the kernel ridge regression formula to make predictions on the test dataset. Since the prediction for each sample is a vector in $\mathbb{R}^{10}$, we take the index of the maximum coordinate as the predicted class label. No regularization ($\lambda=0$) is used unless noted otherwise.

The matrix inversion and matrix multiplication required for kernel ridge regression are computed with GPU acceleration.

\section{Confidence Interval of the Experimental Observations}\label{app:conf}
\begin{figure}
\centering
\includegraphics[width=\linewidth]{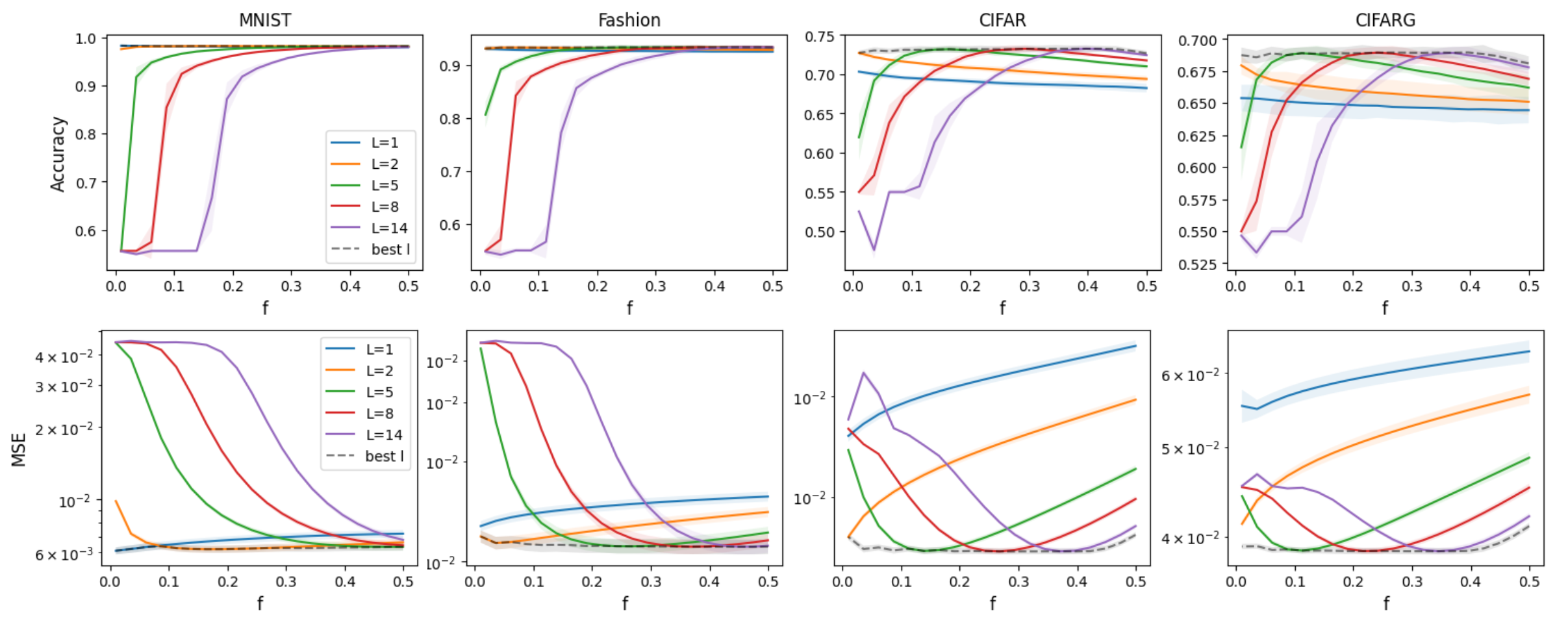}
\caption{The slices of the heatmap plots of Figure \ref{infinite} $P=3982$ at different depths. In the figure, the mean and the $95\%$ confidence interval based on 10 trials with different training sets are shown. The best generalization error and accuracy are shown with the black dashed line.}
\label{fig:confi}
\vspace*{-\intextsep}
\end{figure}

An example figure showing the slice of $fL$-plane with the confidence intervals is shown in Figure \ref{fig:confi}. This shows that the variation in the performance is insensitive to the choice of the random training samples. 

\section{Sparse and Shallow Networks are Comparable Dense and Deep Networks: Additional Results}\label{app:table}

See table \ref{supp_table}.

\begin{table}[]
    \caption{ For each training data, e.g. MNIST with training size $P$ 100, we show in the table, at which depth $L$ the dense model $f=0.5$ performed the best (compared to other depths with $f=0.5$). This depth $L$ and the generalization accuracy at this depth are shown in the first column "Dense - best acc.". We then find a sparse model with comparable performance, and show its $L$, $f$, and generalization accuracy in the second column "Sparse - equiv. acc.". For all datasets, i.e. MNIST and CIFAR10 and all different training set sizes that we tested, the depth of the sparse model that performed comparably to the dense model is at least half of that of the dense counterpart (compare $L$ under the columns "Dense - best acc." and  "Sparse - equiv. acc.") We check the performance of the dense model at the depth at which the sparse model performed comparably to the best dense model. This is to quantify the performance gain we get from sparsity at that given depth (compare the columns "Sparse - equiv. acc." and "Dense - same L"). The table shows there is always a performance gain from sparsity, and this gain is greater when $P$ is smaller or when the task is harder, e.g. CIFAR10. For each setup, we did 10 trials with randomly sampled training sets. We show the standard deviation of accuracy with the $\pm$ notation.}
    \centering
    \begin{tabular}{l c c c c}
    \toprule 
    Dataset: $P$ &  & Dense - best acc. & Sparse - equiv. acc. & Dense - same $L$\tabularnewline
    \cmidrule(r){1-5}
    MNIST: 100 & Accuracy & 0.7545$\pm$0.032 & 0.7560$\pm$0.029 & 0.7292$\pm$0.027\tabularnewline
    \cline{2-5} \cline{3-5} \cline{4-5} \cline{5-5} 
     & $L$ & 7 & 1 & 1\tabularnewline
    \cline{2-5} \cline{3-5} \cline{4-5} \cline{5-5} 
     & $f$ & 0.5 & 0.139 & 0.5\tabularnewline
    \hline 
    MNIST: 500 & Accuracy & 0.8989$\pm$0.0052 & 0.8993$\pm$0.0046 & 0.8836$\pm$0.0037\tabularnewline
    \cline{2-5} \cline{3-5} \cline{4-5} \cline{5-5} 
     & $L$ & 6 & 1 & 1\tabularnewline
    \cline{2-5} \cline{3-5} \cline{4-5} \cline{5-5} 
     & $f$ & 0.5 & 0.139 & 0.5\tabularnewline
    \hline 
    MNIST: 1000 & Accuracy & 0.9300$\pm$0.0028 & 0.9303$\pm$0.0025 & 0.9227$\pm$0.0015\tabularnewline
    \cline{2-5} \cline{3-5} \cline{4-5} \cline{5-5} 
     & $L$ & 5 & 1 & 1\tabularnewline
    \cline{2-5} \cline{3-5} \cline{4-5} \cline{5-5} 
     & $f$ & 0.5 & 0.139 & 0.5\tabularnewline
    \hline 
    MNIST: 2000 & Accuracy & 0.9493$\pm$0.0022 & 0.9495$\pm$0.0021 & 0.9456$\pm$0.0021\tabularnewline
    \cline{2-5} \cline{3-5} \cline{4-5} \cline{5-5} 
     & $L$ & 3 & 1 & 1\tabularnewline
    \cline{2-5} \cline{3-5} \cline{4-5} \cline{5-5} 
     & $f$ & 0.5 & 0.113 & 0.5\tabularnewline
    \hline 
    MNIST: 10000 & Accuracy & 0.9748$\pm$0.0014 & 0.9749$\pm$0.0013 & 0.9725$\pm$0.0013\tabularnewline
    \cline{2-5} \cline{3-5} \cline{4-5} \cline{5-5} 
     & $L$ & 3 & 1 & 1\tabularnewline
    \cline{2-5} \cline{3-5} \cline{4-5} \cline{5-5} 
     & $f$ & 0.5 & 0.087 & 0.5\tabularnewline
    \cmidrule(r){1-5}
    CIFAR10: 100 & Accuracy & 0.2428$\pm$0.014 & 0.2429$\pm$0.014 & 0.2321$\pm$0.012\tabularnewline
    \cline{2-5} \cline{3-5} \cline{4-5} \cline{5-5} 
     & $L$ & 18 & 6 & 6\tabularnewline
    \cline{2-5} \cline{3-5} \cline{4-5} \cline{5-5} 
     & $f$ & 0.5 & 0.294 & 0.5\tabularnewline
    \hline 
    CIFAR10: 500 & Accuracy & 0.3468$\pm$0.011 & 0.3473$\pm$0.012 & 0.3018$\pm$0.011\tabularnewline
    \cline{2-5} \cline{3-5} \cline{4-5} \cline{5-5} 
     & $L$ & 18 & 3 & 3\tabularnewline
    \cline{2-5} \cline{3-5} \cline{4-5} \cline{5-5} 
     & $f$ & 0.5 & 0.087 & 0.5\tabularnewline
    \hline 
    CIFAR10: 1000 & Accuracy & 0.3810$\pm$0.0033 & 0.3814$\pm$0.0046 & 0.3160$\pm$0.0060\tabularnewline
    \cline{2-5} \cline{3-5} \cline{4-5} \cline{5-5} 
     & $L$ & 18 & 2 & 2\tabularnewline
    \cline{2-5} \cline{3-5} \cline{4-5} \cline{5-5} 
     & $f$ & 0.5 & 0.01 & 0.5\tabularnewline
    \hline 
    CIFAR10: 2000 & Accuracy & 0.4156$\pm$0.0026 & 0.4163$\pm$0.0037 & 0.3483$\pm$0.0031\tabularnewline
    \cline{2-5} \cline{3-5} \cline{4-5} \cline{5-5} 
     & $L$ & 18 & 2 & 2\tabularnewline
    \cline{2-5} \cline{3-5} \cline{4-5} \cline{5-5} 
     & $f$ & 0.5 & 0.01 & 0.5\tabularnewline
    \hline 
    CIFAR10: 10000 & Accuracy & 0.5016$\pm$0.0055 & 0.5017$\pm$0.0057 & 0.4621$\pm$0.0035\tabularnewline
    \cline{2-5} \cline{3-5} \cline{4-5} \cline{5-5} 
     & $L$ & 18 & 3 & 3\tabularnewline
    \cline{2-5} \cline{3-5} \cline{4-5} \cline{5-5} 
     & $f$ & 0.5 & 0.087 & 0.5\tabularnewline
    \bottomrule 
    \end{tabular}
    \label{supp_table}
\end{table}

\section{Applying the generalization theory to real dataset}\label{app:apply}

We follow the method provided by Canatar et al. for applying this theory to real datasets to make predictions on the generalization error. Assuming the data distribution $p(x)$ is a discrete uniform distribution over both the training and test datasets, we perform eigendecomposition on a $M\times M$ kernel gram matrix. $M$ is the number of samples across training and test datasets. We need to divide the resulting eigenvalues by the number of non-zero eigenvalues $N$, and multiply the eigenvectors by $\sqrt{M}$ to obtain the finite and discrete estimations of $\eta_\rho =\mathcal{O}(1)$ and $\phi_\rho (\mathbf{x}) =\mathcal{O}(1)$ respectively. Assuming $\mathbf{\Phi}$ is a matrix whose columns are the eigenvectors obtained by the eigendecomposition (before the scaling), the target function coefficients are given by the elements of $\mathbf{\bar{v}} = \frac{1}{\sqrt{M}}\mathbf{\Phi}^\top \mathbf{Y}$ $\in \mathbb{R}^{N}$. The vector $\mathbf{Y}\in\mathbb{R}^{M}$ is the target vector that contains both the training and test sets.

Note that we assume there is no noise added to the target function. In this limit, $\alpha=P/N$ that is greater than 1 results in exactly $0$ generalization error \citetSM{canatar2021spectral}. For the NNGP kernel and the dataset we use, we always have $M=N$, so naturally we have $\alpha<1$.

\section{Derivative of $E_g$}\label{app:dEg}

We want to compute the derivative of $E_{g}$ with respect to eigenvalues.
We decompose the derivative as the following.

\begin{equation}
\frac{dE_{g}}{d\eta_{i}}=\bar{v}_{i}^{2}\frac{d}{d\eta_{i}}E_{i}+\sum_{i\neq\rho}\bar{v}_{\rho}^{2}\frac{d}{d\eta_{i}}E_{\rho}
\end{equation}
The problem boils down to solving the derivatives of the modal errors,
$\frac{d}{d\eta_{i}}E_{i}$ and $\frac{d}{d\eta_{i}}E_{\rho}$ for $i\neq\rho$.

First compute $\frac{d}{d\eta_{i}}E_{i}$.

\begin{equation}
\frac{d}{d\eta_{i}}E_{i}=\left(1-\gamma\right)^{-2}\frac{d\gamma}{d\eta_{i}}\left(1+P\eta_{i}\kappa^{-1}\right)^{-2}-2\left(1-\gamma\right)^{-1}\left(1+P\eta_{i}\kappa^{-1}\right)^{-3}\left(P\kappa^{-1}-P\eta_{i}\kappa^{-2}\frac{d\kappa}{d\eta_{i}}\right)
\end{equation}

\begin{multline}
=\left(1-\gamma\right)^{-2}\frac{d\gamma}{d\eta_{i}}\left(1+P\eta_{i}\kappa^{-1}\right)^{-2}-2\left(1-\gamma\right)^{-1}\left(1+P\eta_{i}\kappa^{-1}\right)^{-3}P\kappa^{-1}
\\
+2\left(1-\gamma\right)^{-1}\left(1+P\eta_{i}\kappa^{-1}\right)^{-3}P\eta_{i}\kappa^{-2}\frac{d\kappa}{d\eta_{i}}
\end{multline}

Then compute $\frac{d}{d\eta_{i}}E_{\rho}$.

\begin{equation}
\frac{d}{d\eta_{i}}E_{\rho}=\left(1-\gamma\right)^{-2}\frac{d\gamma}{d\eta_{i}}\left(1+P\eta_{\rho}\kappa^{-1}\right)^{-2}+2\left(1-\gamma\right)^{-1}\left(1+P\eta_{\rho}\kappa^{-1}\right)^{-3}P\eta_{\rho}\kappa^{-2}\frac{d\kappa}{d\eta_{i}}
\end{equation}

This means the derivative of $E_{g}$ is the following.

\begin{multline}
\frac{dE_{g}}{d\eta_{i}}=-\bar{v}_{i}^{2}2\left(1-\gamma\right)^{-1}\left(1+P\eta_{i}\kappa^{-1}\right)^{-3}P\kappa^{-1}+\left(1-\gamma\right)^{-2}\frac{d\gamma}{d\eta_{i}}\sum_{\rho}\bar{v}_{\rho}^{2}\left(1+P\eta_{\rho}\kappa^{-1}\right)^{-2}
\\
+2P\left(1-\gamma\right)^{-1}\kappa^{-2}\frac{d\kappa}{d\eta_{i}}\sum_{\rho}\bar{v}_{\rho}^{2}\left(1+P\eta_{\rho}\kappa^{-1}\right)^{-3}\eta_{\rho}    
\end{multline}

\begin{equation}
=-\bar{v}_{i}^{2}2\left(1-\gamma\right)^{-1}\left(1+P\eta_{i}\kappa^{-1}\right)^{-3}P\kappa^{-1}+\left(1-\gamma\right)^{-2}\frac{d\gamma}{d\eta_{i}}a+2P\left(1-\gamma\right)^{-1}\kappa^{-2}\frac{d\kappa}{d\eta_{i}}b
\end{equation}

where

\begin{equation}
a=\sum_{\rho}\bar{v}_{\rho}^{2}\left(1+P\eta_{\rho}\kappa^{-1}\right)^{-2}
\end{equation}

\begin{equation}
b=\sum_{\rho}\bar{v}_{\rho}^{2}\left(1+P\eta_{\rho}\kappa^{-1}\right)^{-3}\eta_{\rho}
\end{equation}

\begin{equation}
c=\sum_{\rho}\left(\kappa\eta_{\rho}^{-1}+P\right)^{-3}\eta_{\rho}^{-1}
\end{equation}

We now compute $\frac{d\kappa}{d\eta_{i}}$.

\begin{equation}
\kappa=\lambda+\sum_{\rho}\frac{\kappa\eta_{\rho}}{\kappa+P\eta_{\rho}}=\lambda+\sum_{\rho}\left(\eta_{\rho}^{-1}+P\kappa^{-1}\right)^{-1}
\end{equation}

\begin{align}
\frac{d\kappa}{d\eta_{i}} =&\frac{d}{d\eta_{i}}\left(\eta_{i}^{-1}+P\kappa^{-1}\right)^{-1}+\sum_{\rho\neq i}\frac{d}{d\eta_{i}}\left(\eta_{\rho}^{-1}+P\kappa^{-1}\right)^{-1}
\\
=&\left(\eta_{i}^{-1}+P\kappa^{-1}\right)^{-2}\left(\eta_{i}^{-2}+P\kappa^{-2}\frac{d\kappa}{d\eta_{i}}\right)+\sum_{\rho\neq i}\left(\eta_{\rho}^{-1}+P\kappa^{-1}\right)^{-2}\left(P\kappa^{-2}\frac{d\kappa}{d\eta_{i}}\right)
\\
=&\left(1+P\eta_{i}\kappa^{-1}\right)^{-2}+P\frac{d\kappa}{d\eta_{i}}\left(\eta_{i}^{-1}\kappa+P\right)^{-2}+P\frac{d\kappa}{d\eta_{i}}\sum_{\rho\neq i}\left(\eta_{\rho}^{-1}\kappa+P\right)^{-2}
\\
=&\left(1+P\eta_{i}\kappa^{-1}\right)^{-2}+P\frac{d\kappa}{d\eta_{i}}\sum_{\rho}\left(\eta_{\rho}^{-1}\kappa+P\right)^{-2}    
\end{align}

\begin{equation}
\frac{d\kappa}{d\eta_{i}}\left(1-P\sum_{\rho}\left(\eta_{\rho}^{-1}\kappa+P\right)^{-2}\right)=\left(1+P\eta_{i}\kappa^{-1}\right)^{-2}
\end{equation}

\begin{equation}
\frac{d\kappa}{d\eta_{i}}=\left(1-\gamma\right)^{-1}\left(1+P\eta_{i}\kappa^{-1}\right)^{-2}
\end{equation}

We now compute $\frac{d\gamma}{d\eta_{i}}$.
\begin{equation}
\gamma=\sum_{\rho}\frac{P\eta_{\rho}^{2}}{\left(\kappa+P\eta_{\rho}\right)^{2}}=\sum_{\rho}\frac{P}{\left(\kappa\eta_{\rho}^{-1}+P\right)^{2}}    
\end{equation}

\begin{align}
\frac{1}{P}\frac{d\gamma}{d\eta_{i}} =& \frac{d}{d\eta_{i}}\left(\kappa\eta_{i}^{-1}+P\right)^{-2}+\sum_{\rho\neq i}\frac{d}{d\eta_{i}}\left(\kappa\eta_{\rho}^{-1}+P\right)^{-2}    
\\
=& -2\left(\kappa\eta_{i}^{-1}+P\right)^{-3}\left(\frac{d\kappa}{d\eta_{i}}\eta_{i}^{-1}-\kappa\eta_{i}^{-2}\right)-2\frac{d\kappa}{d\eta_{i}}\sum_{\rho\neq i}\left(\kappa\eta_{\rho}^{-1}+P\right)^{-3}\eta_{\rho}^{-1}
\\
=& -2\left(\kappa\eta_{i}^{-1}+P\right)^{-3}\frac{d\kappa}{d\eta_{i}}\eta_{i}^{-1}+2\left(\kappa\eta_{i}^{-1}+P\right)^{-3}\kappa\eta_{i}^{-2}-2\frac{d\kappa}{d\eta_{i}}\sum_{\rho\neq i}\left(\kappa\eta_{\rho}^{-1}+P\right)^{-3}\eta_{\rho}^{-1}
\\
=& 2\left(\kappa\eta_{i}^{-1}+P\right)^{-3}\kappa\eta_{i}^{-2}-2\frac{d\kappa}{d\eta_{i}}\sum_{\rho}\left(\kappa\eta_{\rho}^{-1}+P\right)^{-3}\eta_{\rho}^{-1}
\end{align}

Substituting $\frac{d\gamma}{d\eta_{i}}$ in $\frac{dE_{g}}{d\eta_{i}}$, we have

\begin{multline}
\frac{1}{2}\left(1-\gamma\right)\frac{dE_{g}}{d\eta_{i}}=-\bar{v}_{i}^{2}P\kappa^{-1}\left(1+P\eta_{i}\kappa^{-1}\right)^{-3}\\
+a\kappa^{-2}\eta_{i}\left(1-\gamma\right)^{-1}\left(1+P\eta_{i}\kappa^{-1}\right)^{-3}-ac\left(1-\gamma\right)^{-1}\frac{d\kappa}{d\eta_{i}}+bP\kappa^{-2}\frac{d\kappa}{d\eta_{i}}    
\end{multline}

Substituting $\frac{d\kappa}{d\eta_{i}}$ in $\frac{dE_{g}}{d\eta_{i}}$, we finally have

\begin{equation}
\frac{dE_{g}}{d\eta_{i}}=2\kappa\left(\eta_{i}\kappa E_{g}-\bar{v}_{i}^{2}P\kappa^{2}\right)\left(1-\gamma\right)^{-1}\left(\kappa+P\eta_{i}\right)^{-3}+2\left(bP-cE_{g}\kappa^{2}\right)\left(1-\gamma\right)^{-2}\left(\kappa+P\eta_{i}\right)^{-2}
\end{equation}.

\section{Perturbation Analysis on $E_\rho$}\label{app:perturb}
We first compute the Jacobian of $E_\rho$'s with respect to $\eta_\rho$'s evaluated at the flat spectrum, i.e. where all $\eta_\rho$'s are the same. The eigenvalues are denoted $\eta$ in this section from here on. In this scenario, the Jacobian simplifies to

\begin{equation}
    \mathbf{J}(\alpha) = 2\left(1-\alpha\right)\alpha\frac{1}{\eta}\mathbf{M}
\end{equation}
\begin{equation}
    \mathbf{M}_{ij} = -\delta_{ij} + \frac{1}{N}(1-\delta_{ij})
\end{equation}
where $N$ is the number of non-zero eigenvalues, and $\alpha=\frac{P}{N}$ is a ratio of the training set to the number of non-zero eigenvalues. We assume $P\rightarrow \infty$ and $N\rightarrow \infty$, but $\alpha\sim\mathcal{O}(1)$. $\mathbf{M}\in \mathbb{R}^{N\times N}$ is a matrix with $-1$'s on the diagonal and $\frac{1}{N}$'s off-diagonal. To see how $E_\rho$'s change with a perturbation in the spectrum, we dot $\mathbf{J}(\alpha)$ with an eigenvalue perturbation vector $\mathbf{\nabla r}=\left[\nabla \eta_0, \ldots, \nabla \eta_\rho, \ldots, \nabla \eta_{N-1} \right]$, where $\nabla \eta_\rho$'s are the individual perturbations in the eigenvalues. Since the perturbed spectrum is ordered from the largest eigenvalue to the smallest eigenvalue, i.e. $\eta + \nabla \eta_\rho > \eta + \nabla \eta_{\rho'}$, $\nabla \eta_\rho$ should be ordered in a way such that for $\rho<\rho'$, $\nabla \eta_\rho > \nabla \eta_{\rho'}$ in the vector $\nabla \mathbf{r}$. The dot product gives the following result on the change in $E_\rho$.

\begin{equation}
    \nabla E_\rho = -2\left(1-\alpha\right)\alpha \frac{1}{\eta} \left( \nabla \eta_\rho - \langle{\nabla \eta_\rho}\rangle \right) 
\end{equation}

where  $\langle{\nabla \eta_\rho}\rangle = \frac{1}{N} \sum_\rho \nabla \eta_\rho$ is a mean value of the perturbations. Intuitively, the modal error perturbation $\nabla E_\rho $ is a sign-flipped version of the zero-meaned $\nabla \eta_\rho$.

\begin{figure*}[ht]
\begin{center}
\centerline{\includegraphics[width=0.7\textwidth]{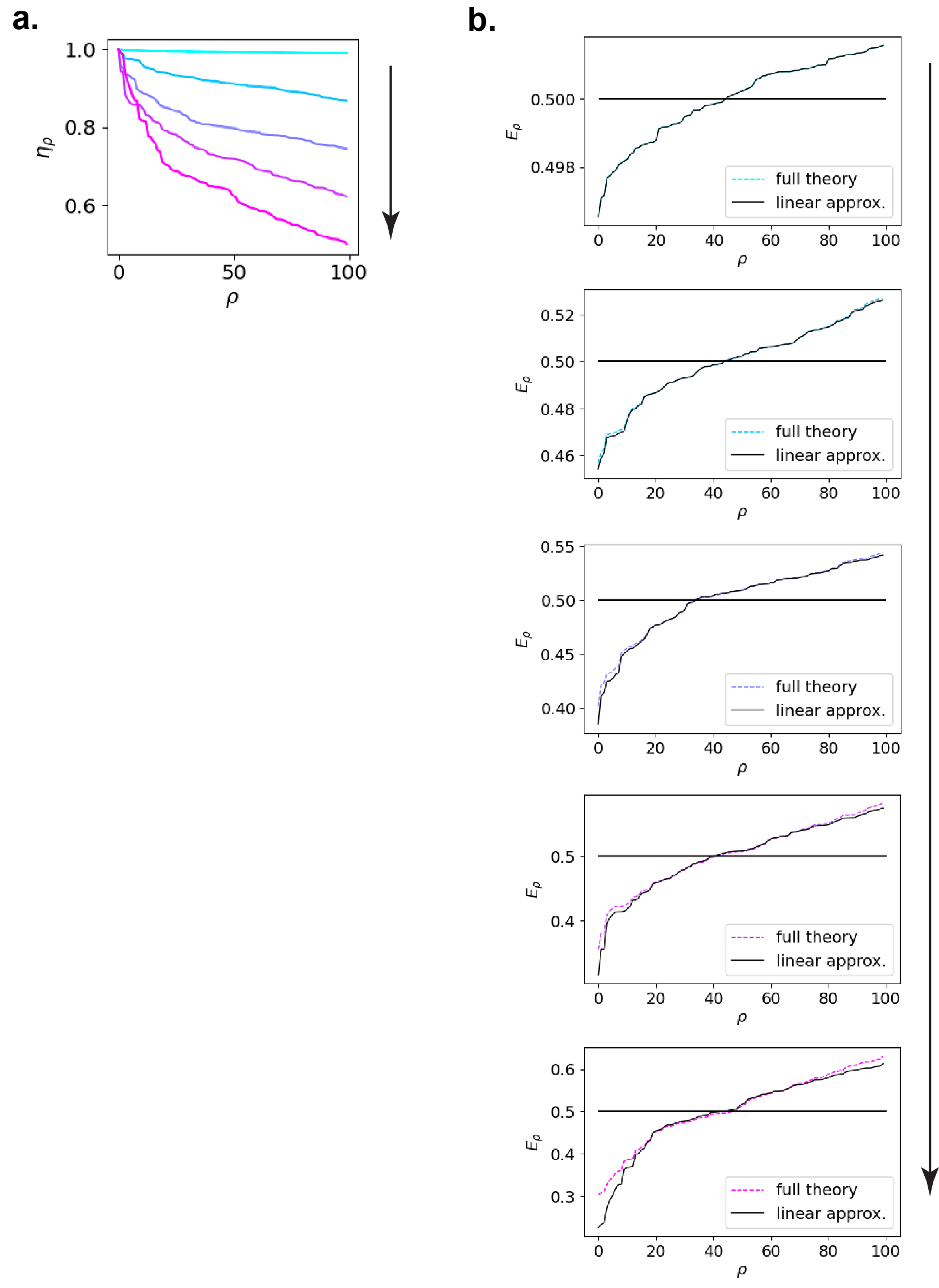}}
\end{center}
\caption{Comparison between the modal error spectrum estimate from the first-order perturbation theory and that of the full theory. (a) The eigenspectrum perturbations from the flat spectrum. (b) The comparisons are ordered from top to bottom in the order of increasing perturbation magnitude. We see that the first-order perturbation accurately predicts the new modal spectrum at high accuracy (top), but it diverges from the full-theoretical result as the perturbation magnitude increases (bottom).}
\label{supp_perval}
\end{figure*}

In Figure \ref{supp_perval}, we compare a modal error spectrum estimated with our first-order perturbation theory to that of the full theory. We see that the first-order perturbation theory accurately predicts the change in the modal error spectrum when the perturbation is small.

\section{Invariance of the Generalization Error to the Change in a Single Eigenvalue}\label{app:invariance}

In this section, we analyze the effect of the offset of the kernel
on the generalization performance. Taking offset is defined as adding
a constant value to a kernel function $K(x,x')+b$. The posterior
output $\mathbf{\hat{y}}$ of kernel ridge regression is given by

\begin{equation}
\mathbf{\hat{y}}=\mathbf{K}'\left(\mathbf{K}+\lambda\mathbf{I}\right)^{-1}\mathbf{y}
\end{equation}

where $\mathbf{y}\in\mathbb{R}^{P}$ is a vector of training labels,
$\mathbf{K}\in\mathbb{R}^{P\times P}$ is a Gram matrix representing
the kernel values amongst the training data, and $\mathbf{K}'\in\mathbb{R}^{M\times P}$
is a Gram matrix representing the kernel values between the test and
training data. Now we introduce the offset to the kernel and the resulting
regression output is

\begin{equation}
\mathbf{\hat{y}}_{b}=\left(\mathbf{K}'+b\mathbf{1}_{M}\mathbf{1}_{P}^{\top}\right)\left(\mathbf{K}+b\mathbf{1}_{P}\mathbf{1}_{P}^{\top}+\lambda\mathbf{I}\right)^{-1}\mathbf{y}
\end{equation}
where $b\mathbf{1}_{M}\mathbf{1}_{P}^{\top}$ is a $M\times P$ matrix
of constant value $b$.

We claim that if one of the eigenvectors of $\mathbf{K}$ is an uniform vector $\phi_{0}=\left[\frac{1}{\sqrt{N}},\ldots,\frac{1}{\sqrt{N}}\right]$,
and the target function is zero-mean (hence $\mathbf{1}_{P}^{\top}\mathbf{y}=0$),
then $\mathbf{\hat{y}}=\mathbf{\hat{y}}_{b}$. Notice that all three
terms in the matrix inverse $\mathbf{G}_{b}=\left(\mathbf{K}+b\mathbf{1}_{P}\mathbf{1}_{P}^{\top}+\lambda\mathbf{I}\right)^{-1}$
are simultaneously diagonalizable. Therefore, if the eigenvalues of
$\left(\mathbf{K}+\lambda\mathbf{I}\right)^{-1}$ are $\left\{ \frac{1}{s_{0}+\lambda},\frac{1}{s_{1}+\lambda},\text{\ensuremath{\ldots}},\frac{1}{s_{P-1}+\lambda}\right\} $,
then the eigenvalues of $\mathbf{G}_{b}$ is $\left\{ \frac{1}{s_{0}+Pb+\lambda},\frac{1}{s_{1}+\lambda},\text{\ensuremath{\ldots}},\frac{1}{s_{P-1}+\lambda}\right\} $
($s_{i}$ is an eigenvalue of $\mathbf{K}$). The only difference
in the eigenspectrums is the first eigenvalue that corresponds to
the uniform eigenvector $\phi_{0}$. Therefore we can decompose $\mathbf{G}_{b}$
in the following fashion.

\begin{equation}
\mathbf{G}_{b}=\left(\mathbf{K}+\lambda\mathbf{I}\right)^{-1}+d\mathbf{1}_{P}\mathbf{1}_{P}^{\top}
\end{equation}
This is equivalent to the Woodbury matrix identity. The specific expression
of $d\in\mathbb{R}$ is irrelevant to our purpose. Therefore, the regression
output is

\begin{align}
\mathbf{\hat{y}}_{b}=&\left(\mathbf{K}'+b\mathbf{1}_{M}\mathbf{1}_{P}^{\top}\right)\left(\left(\mathbf{K}+\lambda\mathbf{I}\right)^{-1}+d\mathbf{1}_{P}\mathbf{1}_{P}^{\top}\right)\mathbf{y}
\\
=&\left(\mathbf{K}'+b\mathbf{1}_{M}\mathbf{1}_{P}^{\top}\right)\left(\left(\mathbf{K}+\lambda\mathbf{I}\right)^{-1}\mathbf{y}+d\mathbf{1}_{P}\mathbf{1}_{P}^{\top}\mathbf{y}\right)
\\
=&\left(\mathbf{K}'+b\mathbf{1}_{M}\mathbf{1}_{P}^{\top}\right)\left(\mathbf{K}+\lambda\mathbf{I}\right)^{-1}\mathbf{y}
\\
=&\mathbf{K}'\left(\mathbf{K}+\lambda\mathbf{I}\right)^{-1}\mathbf{y}+b\mathbf{1}_{M}\mathbf{1}_{P}^{\top}\left(\mathbf{K}+\lambda\mathbf{I}\right)^{-1}\mathbf{y}
\\
=&\mathbf{K}'\left(\mathbf{K}+\lambda\mathbf{I}\right)^{-1}\mathbf{y}
\\
=&\mathbf{\hat{y}}
\end{align}
The third equality is due to $\mathbf{1}_{P}^{\top}\mathbf{y}=0$,
and the fifth equality is due to the fact that $\mathbf{1}_{P}^{\top}\left(\mathbf{K}+\lambda\mathbf{I}\right)^{-1}\propto\mathbf{1}_{P}^{\top}$
since $\mathbf{1}_{P}^{\top}$ is an eigenvector of $\left(\mathbf{K}+\lambda\mathbf{I}\right)^{-1}$,
and therefore

\begin{equation}
\mathbf{1}_{P}^{\top}\left(\mathbf{K}+\lambda\mathbf{I}\right)^{-1}\mathbf{y}\propto\mathbf{1}_{P}^{\top}\mathbf{y}=0
\end{equation}
. Hence we conclude that if an eigenvector of $\mathbf{K}$ is an
uniform vector and the target function is zero-mean, the offset of
the kernel does not affect the kernel regression prediction, and therefore
does not affect the generalization performance. More generally, an additive perturbation $b\phi_i \phi_i^\top $ to a kernel Gram matrix in an eigenvector direction $\phi_i$ that the target function is orthogonal to $\phi_i^\top\mathbf{y}=0$ does not influence the prediction and the generalization performance.

Here we show the alternative proof using the generalization theory.
The generalization error for a kernel that has a constant uniform
vector is expressed as

\begin{equation}
E_{g}=\frac{1}{1-\gamma}\sum_{\rho=1}^{N}\frac{\kappa^{2}\bar{v}_{\rho}^{2}+P\sigma^{2}\eta_{\rho}^{2}}{\left(\kappa+P\eta_{\rho}\right)^{2}}+\frac{1+\gamma}{1-\gamma}\kappa^{2}\frac{\bar{v}_{0}^{2}}{\left(\kappa+2P\eta_{0}\right)^{2}}
\end{equation}

\begin{equation}
\kappa=\lambda+\sum_{\rho=1}^{N}\frac{\kappa\eta_{\rho}}{P\eta_{\rho}+\kappa}
\end{equation}

\begin{equation}
\gamma=\sum_{\rho=1}^{N}\frac{P\eta_{\rho}^{2}}{\left(P\eta_{\rho}+\kappa\right)^{2}}
\end{equation}
where $\eta_{0}$ is the eigenvalue (from Mercer decomposition) that
corresponds to the constant eigenfunction $\phi_{0}(\cdot)$\citetSM{canatar2021spectral}. When
the target function is zero-mean, $\bar{v}_{0}=0$. Therefore, regardless
of the value of the $\eta_{0}$, which is the only eigenvalue that
changes with the offset to the kernel, the second term of $E_{g}$
is $0$, if the target function is zero-mean. This means that if
the target function is zero-mean, the offset to the kernel does not affect
the generalization error.

\section{Theory vs. Experiment on the Real-life Datasets}\label{app:real}
\begin{figure*}[ht]
\begin{center}
\centerline{\includegraphics[width=0.9\textwidth]{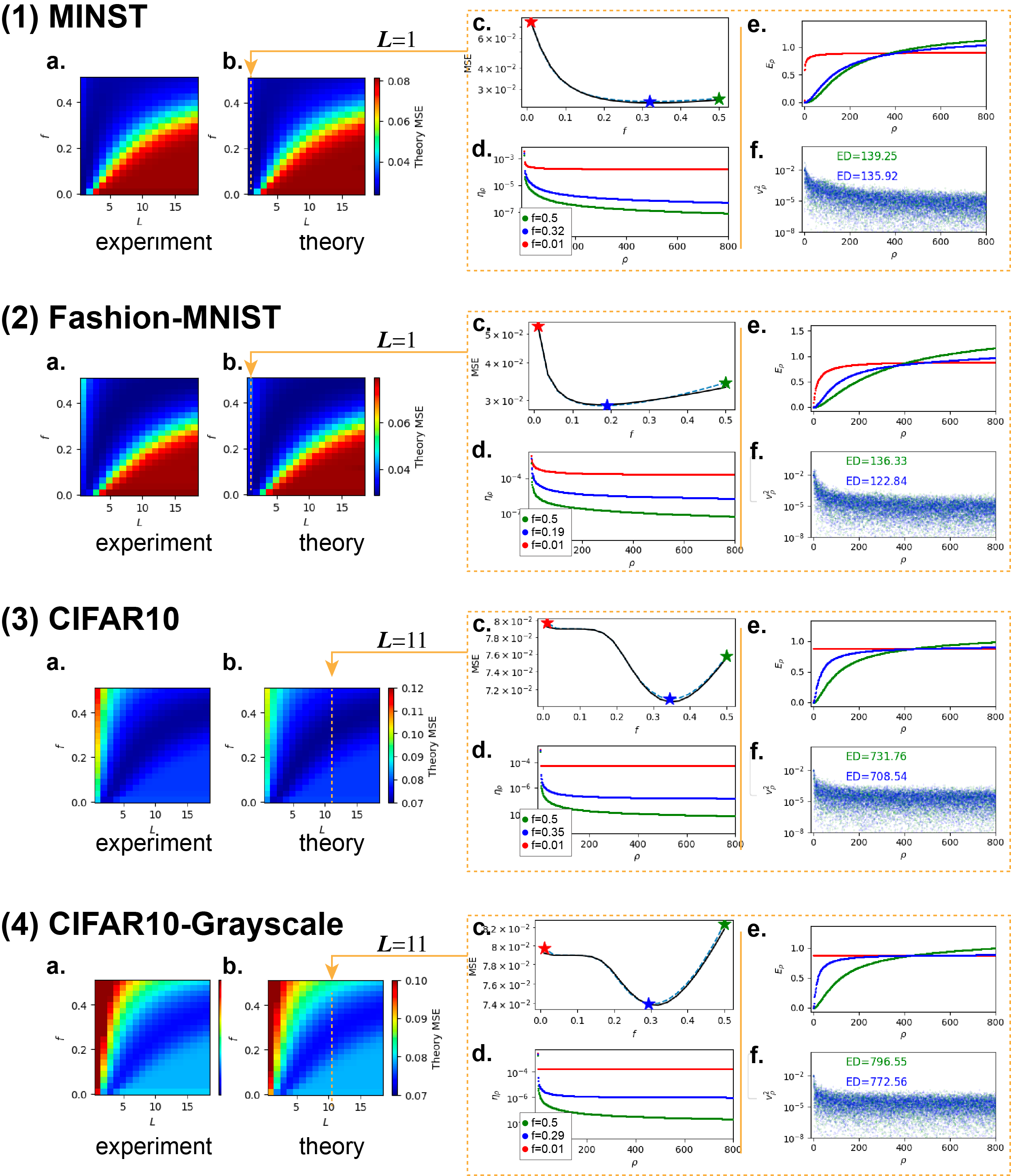}}
\end{center}
\caption{Theoretical analysis of the generalization error over the real-life datasets of $P=362$. (a) The Experimental result on the generalization errors over the circulant dataset over the sparsity ($f$) and depth ($L$). (b) Theoretical predictions of the generalization errors. (c) The generalization error (experimental: blue dotted line, theoretical prediction: black solid line) of the sparse kernels with the specified depth $L$. The kernel with the highest, lowest, and intermediate generalization errors are indicated with red, blue, and green stars respectively; (d) the eigenspectrums (normalized by the second eigenvalues) of the kernels corresponding to the three cases marked in (c). The first 800 eigenvalues are shown; (e) The modal errors $E_\rho$  corresponding to the three cases marked in (c); (f) The target function power $\bar{v}^2_\rho$ spectrum. The effective dimensionality (ED) of the target function power spectrums are indicated in the figure.}
\label{supp_real_spec}
\end{figure*}

In Figure \ref{circ}, we presented the experimental observation of the generalization errors and theoretical predictions on the circulant dataset. We then analyze the shape of the eigenspectrums and the modal spectrums to see what contributed to decreasing or increasing the generalization error. Here, we show the same result on the MNIST, Fashion-MNIST, CIFAR10 and CIFAR10-Grayscale datasets (Figure \ref{supp_real_spec}). Just as in the circulant dataset, we see that the moderately steep eigenspectrum performs the best. In the case of the real-life datasets, the reason the steep eigenspectrum underperforms is because of the large modal errors in modes that correspond to the low eigenvalues, and there are some significant amount of target function powers in those eigenmodes. In the plots, the first 800 eigenmodes are shown. As a reference, there are total ~3000 eigenmodes. We see that the target power spectrum does not vary significantly, as visually shown in the f plots and in the indicated ED values of the target power spectrum.

\section{Comparison to Task-model-alignment in Terms of the Target Power spectrum}\label{app:tma}

In \citetSM{canatar2021spectral}, they present the task-model-alignment as a normalized cumulative sum of the target function powerspectrum.
\begin{equation}
    C(\rho) = \frac{\sum_{i=0}^{\rho-1} \bar{v_i}^2}{\sum_{i=0}^{N-1} \bar{v_i}^2}
\end{equation}

\begin{figure*}[ht]
\begin{center}
\centerline{\includegraphics[width=0.4\textwidth]{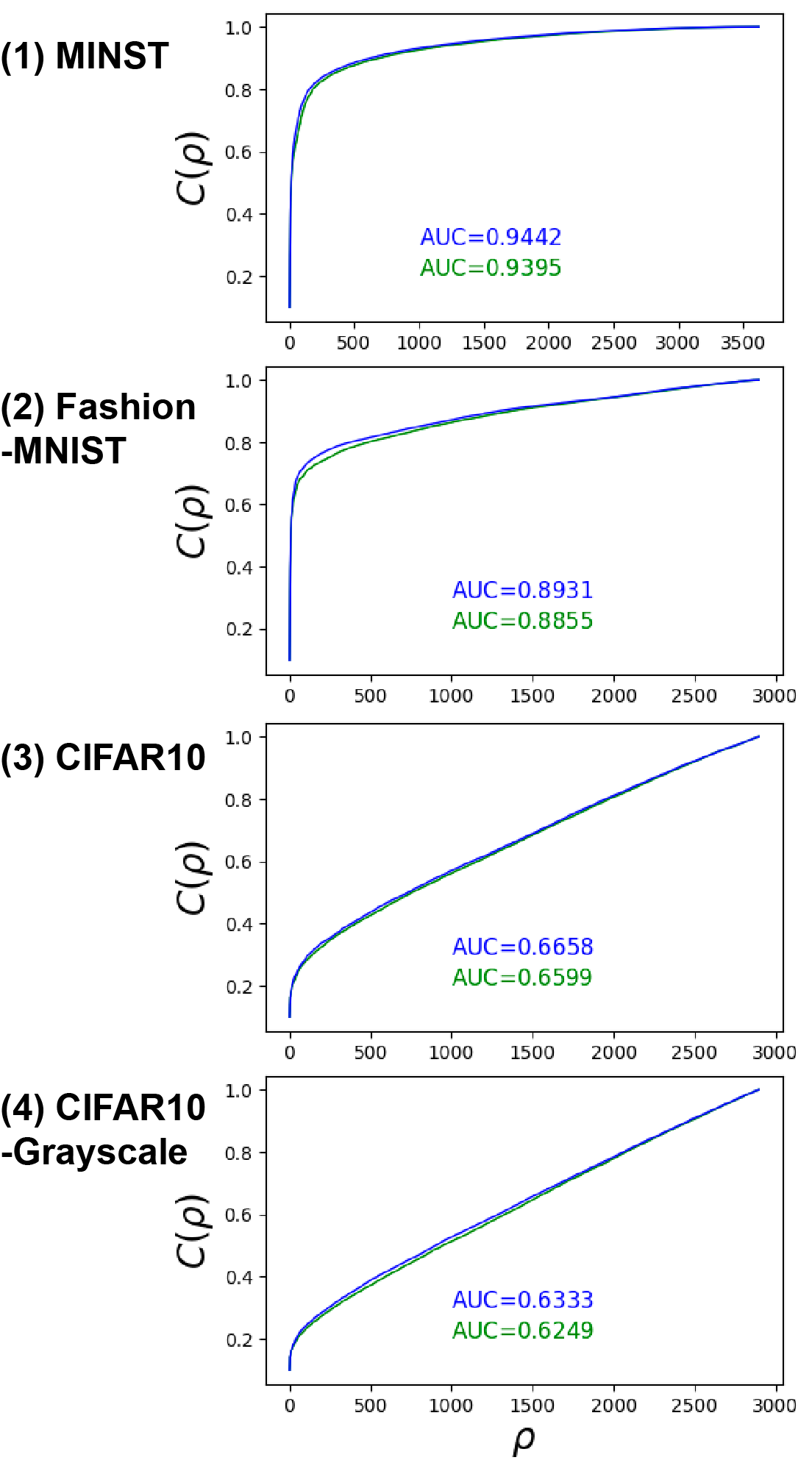}}
\end{center}
\caption{Task-model alignment in the sense of target function power spectrum of the model-data presented in Figure \ref{supp_real_spec}. As shown by the curves and the area under the curves (AUC), the alignment differences are small and hard to tell.}
\label{supp_tma}
\end{figure*}

The faster the rise of the cumulative power, the more aligned the kernel is to the target function, and therefore the generalization performance is better. However, it is challenging to compare the kernels using this metric when the kernel Gram matrix eigenfunctions do not change much between the models. In this circulant case, the eigenfunctions do not change at all, so the target function power spectrum is identical between any kernels. In this case, the comparison using the task-model alignment in the sense of the target function power spectrum fails. Here, we show that we see similar phenomena the real-life datasets (Figure \ref{supp_tma}). The models shown with green and blue colors correspond to the models shown in Figure \ref{supp_real_spec}. 
Here we compute the area under the curve (AUC) of the cumulative sum curve to compare how fast these curves rise. Higher AUC may indicate better-aligned model. We observe that while the higher-performing models (blue ones in Figure \ref{supp_tma}) do have higher AUC than the lower-performing models (green ones in Figure \ref{supp_tma}), the difference is very small. It is also qualitatively hard to tell which model has better alignment just by inspecting the cumulative spectrum. This highlights that our theoretical result on the explicit relationship between the eigenspectrum and the modal error spectrum can complement the task-model alignment presented in \citetSM{canatar2021spectral}.

\bibliographystyleSM{unsrtnat}
\bibliographySM{biblio}

\end{document}